\documentclass[letterpaper]{article}
\usepackage[preprint]{aaai2027}
\usepackage[hyphens]{url}
\usepackage{graphicx}
\usepackage{booktabs}
\usepackage{amsfonts}
\usepackage{amsmath,amssymb}
\usepackage{nicefrac}
\usepackage{xcolor}
\usepackage{tabularx}
\usepackage{multirow}
\usepackage{enumitem}
\usepackage{placeins}
\usepackage{float}

\def\UrlFont{\rm}
\usepackage{natbib}
\usepackage{caption}
\newcolumntype{Y}{>{\centering\arraybackslash}X}
\newcolumntype{L}{>{\raggedright\arraybackslash}X}

\newcommand{\benchmark}{\textsc{SIGNPOST-Bench}}

\title{\benchmark: Benchmarking Text--Vision Conflict Resolution\\in Multimodal Large Language Models}

\author{
    Sirun Li\textsuperscript{\rm 1},
    Minghao Liu\textsuperscript{\rm 2},
    Ling Dai\textsuperscript{\rm 1},
    Yong Li\textsuperscript{\rm 3},
    Haoxin Lyu\textsuperscript{\rm 1},
    Junting Zhou\textsuperscript{\rm 1},
    Fan Zhang\textsuperscript{\rm 1}\corresponding
}
\affiliations{
    \textsuperscript{\rm 1}Peking University, Beijing, China\\
    \textsuperscript{\rm 2}2077AI\\
    \textsuperscript{\rm 3}The Hong Kong University of Science and Technology, Hong Kong, China\\
    fanzhanggis@pku.edu.cn
}

\begin{document}

\maketitle

\begin{abstract}
Multimodal large language models (MLLMs) make grounded predictions in real-world scenes by combining visual and textual cues, yet existing benchmarks rarely reveal how they arbitrate between these evidence sources when they conflict. We introduce \benchmark, a controlled counterfactual benchmark for evaluating text--vision conflict resolution. Each source image is transformed into a counterfactual quintuplet of Original, Blank, Similar, Random, and Adversarial variants. Synthetic, localized scene-text interventions are designed to preserve non-textual content, enabling paired measurements of changes in localization performance and directed shifts toward geographic targets introduced by conflicting text. \benchmark{} contains 5{,}111 counterfactual groups and 25{,}555 image variants from four datasets. We evaluate 20 MLLMs from seven providers. Compared with Original images, Adversarial variants raise median localization error from 282\,km to 1{,}347\,km, a 4.8-fold increase. Among geocodable adversarial samples, 6.5--20.1\% of predictions lie less than 50\,km from the injected target across models, and every evaluated model exhibits a positive mean paired reduction in target distance from Blank to Adversarial. Compatible, unrelated, and conflicting text replacements produce distinct effects on model predictions, while clean-input localization performance does not fully predict robustness to conflicting text. These results establish visual geolocation as a continuous diagnostic of scene-text arbitration and provide a controlled framework for evaluating how MLLMs resolve conflicting multimodal evidence.
\end{abstract}

\section{Introduction}

Multimodal large language models (MLLMs) increasingly use both visual context and scene text to make grounded predictions in real-world scenes~\cite{georeasoner,vlm_geoguessr_masters}. When these signals agree, their joint use can improve recognition and reasoning; when they conflict, however, a model must decide which source of evidence to trust. This distinction is critical in natural scenes, where architecture, road structure, and terrain may suggest one location while signs, addresses, or storefront names suggest another. Strong performance on aligned inputs therefore provides no direct evidence that a model can reliably choose between competing visual and textual cues.

Existing benchmarks evaluate OCR, visual recognition, multimodal hallucination, typographic robustness, and visual geolocation, but most report only the final prediction without showing which source of evidence drives it. Recent studies have begun to examine this problem more directly. ConText-VQA tests how misleading textual prompts accompanying a visual question affect model answers~\cite{zhang2026contextvqa}. RIO-Bench uses same-scene counterfactuals to evaluate when models should read or ignore in-image text~\cite{waseda2025read}. However, both evaluate conflict through discrete task outcomes. They do not compare compatible, unrelated, and conflicting scene-text replacements within the same scene, nor do they measure the magnitude and direction of prediction shifts in a shared output space.

We use visual geolocation to address this gap because visual scene structure and scene text can independently suggest locations within a shared coordinate space, allowing us to measure both localization degradation and shifts toward an injected geographic target. By changing scene text through localized edits designed to preserve the surrounding non-textual content, we distinguish generic localization degradation from directed shifts toward geographic targets introduced by conflicting text.

We introduce \benchmark{} (\textbf{S}cene \textbf{I}mage \textbf{G}eo-localization with \textbf{N}oisy \textbf{P}erturbation on \textbf{O}bserved \textbf{S}ign \textbf{T}ext), a controlled counterfactual benchmark for evaluating text--vision conflict resolution. Each source image is transformed into a counterfactual quintuplet of Original, Blank, Similar, Random, and Adversarial variants through synthetic, localized scene-text interventions. Together, these five conditions enable within-scene comparisons of native text, text removal, compatible replacement, unrelated replacement, and conflicting replacement. \benchmark{} contains 5{,}111 counterfactual groups and 25{,}555 image variants from four datasets. We evaluate 20 MLLMs from seven providers across all five conditions.

\begin{figure*}[t]
  \centering
  \includegraphics[width=\linewidth]{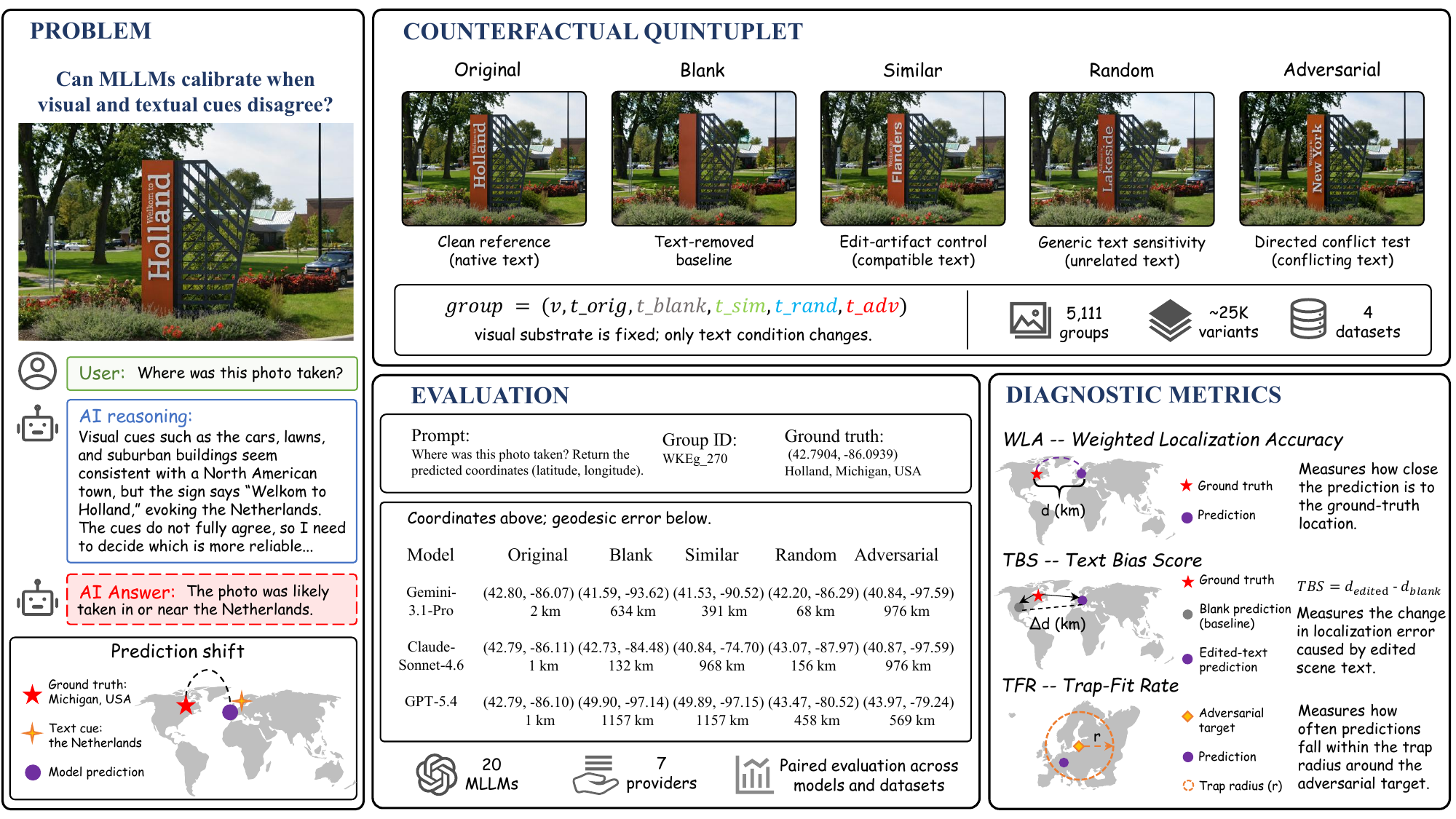}
  \caption{\benchmark{} overview. Left: a scene from Holland, Michigan, illustrates the central problem: native scene text may evoke the Netherlands, whereas the surrounding visual cues support a North American location. Top: synthetic, localized scene-text edits transform each source image into a counterfactual quintuplet of Original, Blank, Similar, Random, and Adversarial variants, providing a native-text reference, a text-removed baseline, a compatible replacement, an unrelated replacement, and a conflicting geographic cue, respectively. Bottom center: the same geolocation prompt is applied to all five variants, enabling within-scene comparison of predictions from 20 MLLMs across seven providers against a shared ground-truth location. Bottom right: Weighted Localization Accuracy (WLA) measures localization performance relative to the ground truth, Text Bias Score (TBS) measures the change in ground-truth geodesic error from Blank to an edited variant, and Trap-Fit Rate (TFR) measures the fraction of Adversarial predictions lying less than 50\,km from the injected geographic target.}
  \label{fig:conflict_mechanism}
\end{figure*}

We study three main questions: How strongly does conflicting scene text reduce localization performance? How do compatible, unrelated, and conflicting text replacements differ in their effects on model predictions? Does conflicting geographic text move predictions toward the injected target rather than simply increasing localization error? We further test whether conflict-aware prompting helps two representative models identify and resolve text--vision conflicts. Figure~\ref{fig:conflict_mechanism} illustrates the benchmark design.

Our contributions are threefold. First, we formulate text--vision conflict resolution in a continuous coordinate space, enabling measurement of both localization degradation and directed shifts toward an injected geographic target. Second, we construct \benchmark{} using a paired five-condition design, with 5{,}111 counterfactual groups and 25{,}555 image variants from four datasets. Third, we evaluate 20 MLLMs from seven providers and show that conflicting scene text substantially degrades localization, shifts predictions toward injected geographic targets on average for every evaluated model, and exposes differences in conflict robustness that clean-input localization performance does not fully predict.

\section{Related Work}
\label{sec:related_work}

Multimodal evaluation has expanded from aggregate accuracy to specific failure modes. General-purpose multimodal suites assess perception, reasoning, and integrated multimodal capabilities~\cite{mme,mmbench,mmvet,seedbench}, while broader model evaluation also targets expert-level knowledge and reasoning across disciplines~\cite{supergpqa}. Targeted multimodal benchmarks examine hallucination, natural adversarial samples, and image adversarial robustness~\cite{pope,hallusionbench,naturalbench,mmrobustness}. Studies identify persistent separation between visual and textual representations~\cite{modality_gap}, while decoding and post-hoc methods mitigate hallucinations associated with weak visual grounding or excessive reliance on language priors~\cite{vcd,opera,woodpecker}. Together, these findings motivate controlled tests of how models arbitrate between conflicting visual and textual evidence.

Scene text is both a visual pattern and a linguistic signal, making it a natural source of cross-modal agreement and conflict. Typographic attacks showed that placing a class name on an image can override visual recognition in CLIP~\cite{typographic_attack}, and subsequent work found systematic task competition among scene-text, object, and action recognition~\cite{task_bias_vlm}. Later studies examined how font, color, position, and semantic relation modulate the effect~\cite{typographic_deceptions}, alongside prompt-based mitigation strategies~\cite{azuma2023defense}. Scene-coherent approaches such as SceneTAP~\cite{scenecoherent_typo} and in-the-wild studies such as SCAM~\cite{scam} further demonstrate that visually integrated text can influence multimodal decisions beyond simple overlaid-word attacks. ConText-VQA evaluates how misleading textual prompts affect answers to visual questions~\cite{zhang2026contextvqa}, while RIO-Bench uses same-scene counterfactuals to test when models should read or ignore in-image text~\cite{waseda2025read}. These studies primarily use discrete classification or VQA outcomes, leaving the directional influence of scene text in a shared geographic output space unmeasured.

Visual geolocation provides a natural setting for such analysis because its outputs are continuous and its errors retain geographic meaning. Retrieval-based and learned approaches, including IM2GPS~\cite{im2gps}, PlaNet~\cite{planet}, GeoCLIP~\cite{geoclip}, StreetCLIP~\cite{streetclip}, and PIGEON~\cite{pigeon}, map visual appearance to geographic locations and evaluate predictions by geodesic distance. Street-view representation learning further uses spatial and temporal metadata to model stable built-environment cues, dynamic content, and neighborhood context~\cite{li2026streetview}. These methods treat each image as a single visual input and therefore do not separate the contribution of readable scene text from the surrounding visual context.

Recent work extends visual geolocation and urban visual reasoning to MLLMs. Models can infer locations from architecture, scripts, signs, and other contextual cues~\cite{georeasoner,vlm_geoguessr_masters}, while VLMs have also been benchmarked on remote-sensing geospatial tasks~\cite{geobenchvlm}. Unified Urban Tuning improves satellite and street-view reasoning across views, cities, and tasks~\cite{li2026uut}, while agentic systems such as SpotAgent and REVERSE incorporate external search and evidence verification~\cite{jia2026spotagent,li2026reverse}. These approaches aim to improve geolocation accuracy or evidence acquisition under naturally observed inputs. In contrast, \benchmark{} applies localized interventions designed to preserve surrounding non-textual content across five matched variants of each source scene---Original, Blank, Similar, Random, and Adversarial---and measures both the magnitude and geographic direction of text-induced prediction shifts.

\section{Benchmark Formulation}

\subsection{Task Definition}

We formulate text--vision conflict resolution through visual geolocation. Given an image $I$, a model $f:\mathcal{I}\rightarrow\mathbb{R}^{2}$ predicts coordinates $(\hat{\mathrm{lat}},\hat{\mathrm{lon}})$. Let $y_{\mathrm{gt}}$ denote the ground-truth coordinates and $\mathcal{D}(\cdot,\cdot)$ the haversine geodesic distance, so the localization error is $\mathcal{D}(f(I),y_{\mathrm{gt}})$.

The experimental unit is a \textbf{counterfactual group} derived from one source scene. Each group contains its ground truth, one or more selected \textbf{scene-text spans}, and five matched images that collectively form a \textbf{counterfactual quintuplet}. A scene-text span is a localized text region selected for removal or replacement. Applying the same model to all five images enables paired, within-scene measurement of text-induced prediction changes.

\subsection{Counterfactual Quintuplet}

The quintuplet comprises Original, Blank, Similar, Random, and Adversarial images (Figure~\ref{fig:conflict_mechanism}). \textbf{Original} is unmodified. \textbf{Blank} removes the selected scene-text spans and serves as the text-ablated reference. \textbf{Similar} replaces them with alternatives compatible with the ground-truth geographic context or language, without requiring literal equivalence to the original text. \textbf{Random} introduces unrelated readable text without a designated geographic target. \textbf{Adversarial} injects a geographically conflicting cue; when geocodable, it defines an injected target $y_{\mathrm{trap}}$.

Comparisons with the corresponding Blank image measure the effects of compatible, unrelated, and conflicting text within the same scene. For geocodable Adversarial images, the shared coordinate space additionally reveals whether predictions shift toward the injected target. Section~\ref{sec:metrics} formalizes these diagnostics.

\subsection{Scene-Text Coupling Taxonomy}

We classify native scene text by its geographic identifiability in the context of the source scene, rather than by fixed lexical categories. \textbf{T1 (Portable)} covers text with little geographic specificity, such as global brands and generic warnings. \textbf{T2 (Cultural)} narrows the location to a language or cultural region without identifying a particular place, as with local scripts or common transit labels. \textbf{T3 (Geo-Specific)} directly identifies a place or distinctive local entity, including street names, addresses, postal codes, district names, and uniquely identifiable businesses. A business name may therefore be T2 or T3 depending on its geographic specificity. The taxonomy supports stratified analysis of text influence; full definitions appear in Appendix~\ref{appendix:tier_stratification} and Table~\ref{tab:coupling_taxonomy}.

\section{Benchmark Construction}

\subsection{Source Collection and Text Screening}

Images come from four complementary sources. IM2GPS3K and YFCC4K~\cite{vo2017revisiting} provide geotagged web photographs with varied viewpoints; YFCC4K is sampled from YFCC100M~\cite{yfcc100m}. GoogleSV and BaiduSV provide international and Chinese street-view imagery. The street-view sources are geographically sampled, after which EasyOCR~\cite{easyocr} identifies candidate scene-text spans in all four sources. Appendix~\ref{appendix:construction_details} reports the sampling, projection, and OCR settings; Figure~\ref{fig:dataset_samples} shows representative images.

\subsection{Counterfactual Generation Pipeline}

Selected images enter a three-stage pipeline for replacement generation, target geocoding, and localized synthesis.

\noindent\textbf{Text replacement generation.} Gemini-3.1-Flash-Lite~\cite{gemini31flashlite} selects up to three informative and editable scene-text spans per image and generates one Similar, Random, and Adversarial replacement for each. When available, source-location metadata are included to ensure that Adversarial text conflicts with the ground truth; these data are used only for construction and are never shown to evaluated models. Replacement constraints and the full prompt appear in Appendices~\ref{appendix:construction_details} and~\ref{appendix:attack_prompt}.

\noindent\textbf{Adversarial target geocoding.} We query place names, landmarks, or addresses in Adversarial text through the Nominatim OpenStreetMap API~\cite{nominatim} and record the top-ranked valid result as the injected target. Groups without a valid target remain in the benchmark but are excluded from TFR and TDR. Table~\ref{tab:tfr_denominator} reports geocoding coverage.

\noindent\textbf{Counterfactual synthesis.} The Qwen-Image-Edit-2509 checkpoint~\cite{qwenimage} removes the selected spans for Blank or renders the corresponding replacements for Similar, Random, and Adversarial. Each edited image is generated independently from Original through localized operations designed to preserve surrounding non-textual content. Appendix~\ref{appendix:construction_details} gives the checkpoint identifier and synthesis settings.

\begin{figure*}[t]
  \centering
  \includegraphics[width=\linewidth]{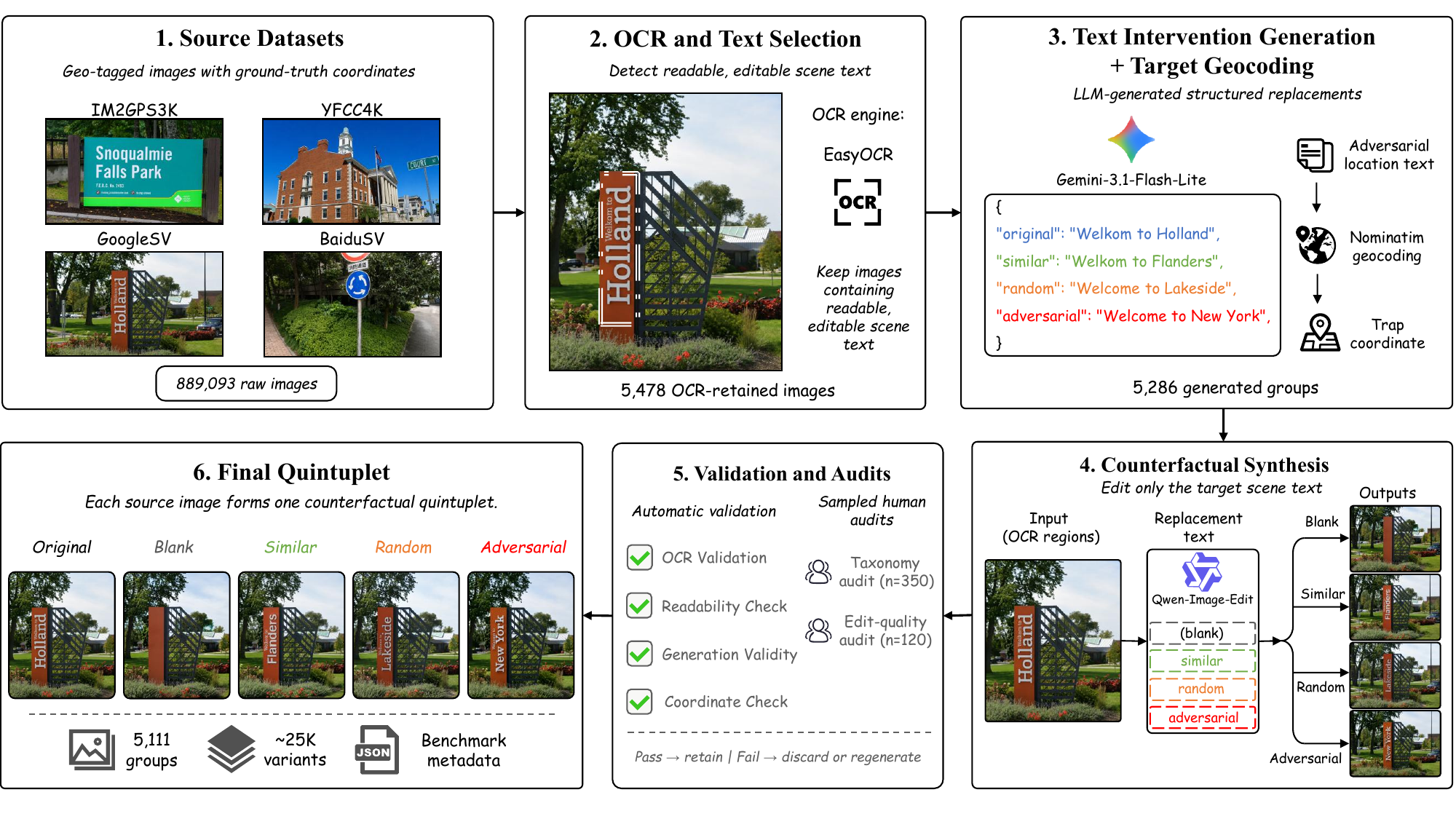}
  \caption{\benchmark{} construction pipeline. Images from four sources are screened for editable scene-text spans. An MLLM generates three replacement types, and localized operations designed to preserve non-textual content produce the four edited images. Together with Original, they form the quintuplet within each counterfactual group. Sampled human audits assess taxonomy labels and edit quality.}
  \label{fig:pipeline_overview}
\end{figure*}

\subsection{Quality Assurance}
\label{sec:quality_assurance}

Two annotators independently assign tiers to 350 stratified source images, obtaining 83.4\% agreement ($\kappa=0.747$). We also audit 120 generated Similar, Random, and Adversarial images. Rendered-text naturalness is rated from 1 (highly unnatural) to 5 (fully natural); artifact severity and surrounding-context damage use 1 (none) to 5 (severe). The corresponding means are $4.00\pm1.26$, $1.32\pm0.78$, and $1.14\pm0.52$. Readability refers to the rendered replacement text: 105 images (87.5\%) are fully readable and 15 (12.5\%) partially readable, with none unreadable. Appendix~\ref{appendix:construction_details} provides the protocol and rubric.

\subsection{Benchmark Composition}

\benchmark{} contains 5{,}111 counterfactual groups, each with one quintuplet, for 25{,}555 images and 10{,}084 scene-text spans. T1, T2, and T3 contain 347 (6.8\%), 3{,}851 (75.3\%), and 913 (17.9\%) groups, respectively. Table~\ref{tab:dataset_stats_supp} gives the construction-stage breakdown.

\section{Evaluation Protocol}

\subsection{Diagnostic Metrics}
\label{sec:metrics}

We use three primary diagnostic metrics. \textbf{Weighted Localization Accuracy (WLA)} scores each prediction by geodesic error via exponential decay, weighting closer predictions more favorably:
\begin{equation}
  \mathrm{WLA}_{i} = \exp(-\alpha d_{i}),
  \qquad
  d_{i} = \mathcal{D}(f(I_{i}), y_{\mathrm{gt}}^{(i)}),
  \label{eq:wla}
\end{equation}
with $\alpha=0.005$ (WLA decays to 0.5 at 138\,km, below 0.01 at 1{,}000\,km). All WLA values are reported as percentages ($\times100$). \textbf{Text Bias Score (TBS)} measures the paired error shift caused by edited text, using the Blank variant as a reference. A positive TBS means that the edited text made the model's prediction worse compared with the same scene with its text removed:
\begin{equation}
  \begin{aligned}
  \mathrm{TBS}_{i}^{(k)}
  &=
  \mathcal{D}\bigl(f(I_{k}^{(i)}), y_{\mathrm{gt}}^{(i)}\bigr) \\
  &\quad -
  \mathcal{D}\bigl(f(I_{\mathrm{blank}}^{(i)}), y_{\mathrm{gt}}^{(i)}\bigr),\\
  &\quad k \in \{\mathrm{sim}, \mathrm{rnd}, \mathrm{adv}\}.
  \end{aligned}
  \label{eq:tbs}
\end{equation}
Because Blank removes the selected text while retaining the surrounding scene, this paired contrast measures the additional change in localization error associated with the edited textual cue relative to the text-removed baseline. TBS is a change in ground-truth error, not the distance between the Blank and edited predictions. \textbf{Trap-Fit Rate (TFR)} measures whether a prediction lands near the injected target. For dataset $d$,
\begin{equation}
  \mathrm{TFR}_{d} =
  \frac{1}{N_d}
  \sum_{i=1}^{N_d}
  \mathbb{I}\!\left(
    \mathcal{D}\bigl(f(I_{\mathrm{adv}}^{(i)}), y_{\mathrm{trap}}^{(i)}\bigr) < \tau
  \right),
  \label{eq:tfr}
\end{equation}
with $\tau = 50$\,km, computed only for adversarial samples whose injected text resolves to valid coordinates. Table~\ref{tab:tfr_denominator} reports the denominators and geocoding rates. We report the equal-dataset macro-average $\mathrm{TFR}=\frac{1}{4}\sum_{d=1}^{4}\mathrm{TFR}_{d}$. WLA and TBS measure prediction quality and error change; TFR measures target proximity. We also report paired \textbf{Trap Distance Reduction (TDR)}:
\begin{equation}
  \mathrm{TDR}_{i} =
  \mathcal{D}\bigl(f(I_{\mathrm{blank}}^{(i)}), y_{\mathrm{trap}}^{(i)}\bigr)
  -
  \mathcal{D}\bigl(f(I_{\mathrm{adv}}^{(i)}), y_{\mathrm{trap}}^{(i)}\bigr),
  \label{eq:tdr}
\end{equation}
where a positive value indicates that the adversarial prediction moves closer to the trap than its paired Blank prediction. TDR is computed on the same geocodable subset as TFR and macro-averaged equally across datasets.

\subsection{Multimodal Conflict Robustness Score}

We decompose evaluation into two components. The \textbf{Capability Score} $C$ averages Original and Blank WLA, capturing localization capability across the native-text and text-removed reference conditions. The \textbf{Conflict Robustness Score} $R$ combines WLA retention relative to Blank under Random and Adversarial text with normalized penalties for error increase ($q_{\mathrm{TBS}}$) and trap-following rate ($q_{\mathrm{TFR}}$):
\begin{equation}
  \begin{aligned}
  C &= 0.50\,w_{\mathrm{orig}} + 0.50\,w_{\mathrm{blank}},\\
  R &= 0.22\,\rho_{\mathrm{rnd}}+0.44\,\rho_{\mathrm{adv}}
  +0.17\,q_{\mathrm{TBS}}+0.17\,q_{\mathrm{TFR}}.
  \end{aligned}
  \label{eq:robustness}
\end{equation}
where $\rho_{\mathrm{rnd}}$ and $\rho_{\mathrm{adv}}$ are WLA retention rates relative to Blank, and $q_{\mathrm{TBS}}$ and $q_{\mathrm{TFR}}$ are normalized quality terms for which higher is better. Both $C$ and $R$ are defined on $[0,1]$; tables report them as percentages for readability. The integrated \textbf{Multimodal Conflict Robustness Score (MCRS)} is
\begin{equation}
  \mathrm{MCRS}=100\cdot C^{0.40}\cdot R^{0.60},
  \label{eq:mcrs}
\end{equation}
with the larger exponent on $R$ reflecting the benchmark's focus on conflict handling. Appendix~\ref{appendix:mcrs_formula} defines clipping and normalization anchors, while Appendix~\ref{appendix:sensitivity} reports weight and exponent sensitivity analyses with Kendall $\tau\geq0.905$.

\subsection{Models and Evaluation Setup}

We evaluate 20 MLLMs from seven providers (4 Google, 4 OpenAI, 3 Anthropic, 3 Alibaba, 3 Moonshot, 2 ByteDance, 1 xAI) on all five variants of all 5{,}111 counterfactual groups, yielding 511{,}100 model--image evaluations. Table~\ref{tab:model_registration} lists the exact API identifiers. Each model receives the same prompt (Appendix~\ref{appendix:coord_prompt}) requesting direct coordinate estimates without chain-of-thought. Metrics are averaged equally across the four datasets to prevent larger street-view sources from dominating.

\section{Results}

\subsection{MLLM Performance Degrades under Text--Vision Conflict}
\label{sec:overall_conflict}

Table~\ref{tab:main_model_summary} reports the complete five-condition performance of all 20 models. Across models and datasets, MLLMs perform substantially worse when scene text conflicts with the surrounding visual evidence.

\begin{table*}[t]
  \centering
  \small
  \setlength{\tabcolsep}{1.5pt}
  \begin{tabular}{@{}llrrrrrrrrrr@{}}
    \toprule
    & & \multicolumn{5}{c}{\textbf{Weighted Localization Accuracy}} & \multicolumn{2}{c}{\textbf{Diagnostics}} & \multicolumn{3}{c}{\textbf{Summary Scores}} \\
    \cmidrule(lr){3-7}\cmidrule(lr){8-9}\cmidrule(lr){10-12}
    \textbf{Provider} & \textbf{Model} & \textbf{Orig.}$\uparrow$ & \textbf{Blank}$\uparrow$ & \textbf{Similar}$\uparrow$ & \textbf{Random}$\uparrow$ & \textbf{Adv.}$\uparrow$ & \textbf{TBS$_{\mathrm{adv}}$ (km)}$\downarrow$ & \textbf{TFR}$\downarrow$ & $\mathbf{C}\uparrow$ & $\mathbf{R}\uparrow$ & \textbf{MCRS}$\uparrow$ \\
    \midrule
    \multirow{4}{*}{Google}
      & Gemini-3.1-Pro   & \textbf{64.01} & \textbf{50.16} & \textbf{57.15} & \textbf{45.41} & \textbf{44.41} & 879 & 8.0\% & \textbf{57.08} & 84.49 & 72.23 \\
      & Gemini-3-Flash   & 63.94 & 49.23 & 56.62 & 45.21 & 43.80 & \textbf{811} & 6.7\% & 56.58 & \textbf{85.93} & \textbf{72.70} \\
      & Gemini-2.5-Pro   & 58.16 & 45.56 & 53.93 & 41.60 & 40.18 & 888 & \textbf{6.5\%} & 51.86 & 85.10 & 69.80 \\
      & Gemini-2.5-Flash & 53.49 & 43.23 & 50.15 & 36.65 & 36.90 & 1{,}441 & 10.0\% & 48.36 & 77.77 & 64.31 \\
    \addlinespace[2pt]
    \multirow{4}{*}{OpenAI}
      & GPT-5        & 55.73 & 45.67 & 52.55 & 37.56 & 36.95 & 1{,}317 & 11.6\% & 50.70 & 75.30 & 64.28 \\
      & GPT-4o       & 47.01 & 37.80 & 43.20 & 27.85 & 28.07 & 1{,}461 & 11.5\% & 42.40 & 69.74 & 57.16 \\
      & GPT-5.4      & 40.25 & 32.02 & 39.47 & 25.32 & 25.90 & 1{,}558 & 13.8\% & 36.14 & 72.27 & 54.77 \\
      & GPT-4o-mini  & 35.37 & 28.42 & 35.12 & 23.57 & 20.35 & 1{,}346 & 16.7\% & 31.89 & 69.02 & 50.68 \\
    \addlinespace[2pt]
    \multirow{3}{*}{Anthropic}
      & Claude-Sonnet-4.6 & 52.39 & 42.42 & 48.88 & 34.81 & 34.66 & 1{,}535 & 11.8\% & 47.41 & 74.30 & 62.08 \\
      & Claude-Opus-4.6   & 51.06 & 40.85 & 47.66 & 35.20 & 34.47 & 1{,}482 & 11.4\% & 45.95 & 76.84 & 62.56 \\
      & Claude-Haiku-4.5  & \underline{27.54} & \underline{20.67} & \underline{26.78} & \underline{12.94} & \underline{12.77} & 2{,}287 & \underline{20.1\%} & \underline{24.10} & \underline{53.46} & \underline{38.87} \\
    \addlinespace[2pt]
    \multirow{3}{*}{Alibaba}
      & Qwen3-VL-30B  & 38.88 & 34.22 & 38.52 & 29.57 & 29.03 & 1{,}100 & 9.6\% & 36.55 & 80.05 & 58.50 \\
      & Qwen3-VL-Plus & 40.66 & 35.61 & 39.45 & 30.41 & 27.92 & 1{,}438 & 12.6\% & 38.13 & 73.80 & 56.67 \\
      & Qwen3-VL-235B & 42.78 & 34.52 & 41.11 & 26.61 & 25.49 & 1{,}787 & 16.9\% & 38.64 & 66.16 & 53.36 \\
    \addlinespace[2pt]
    \multirow{2}{*}{ByteDance}
      & Seed-2.0-Pro  & 61.03 & 48.99 & 54.69 & 42.69 & 38.95 & 1{,}796 & 15.9\% & 55.01 & 71.21 & 64.23 \\
      & Seed-2.0-Lite & 58.36 & 47.20 & 53.09 & 37.97 & 36.47 & 2{,}039 & 15.7\% & 52.78 & 67.47 & 61.16 \\
    \addlinespace[2pt]
    \multirow{3}{*}{Moonshot}
      & Kimi-K2.5          & 55.73 & 45.31 & 51.61 & 34.59 & 35.29 & 1{,}577 & 13.1\% & 50.52 & 70.57 & 61.74 \\
      & Moonshot-128K-Vision & 28.82 & 22.72 & 29.97 & 14.56 & 14.28 & 2{,}090 & 16.4\% & 25.77 & 56.92 & 41.46 \\
      & Moonshot-32K-Vision  & 28.79 & 22.39 & 30.15 & 14.39 & 14.08 & 2{,}045 & 16.4\% & 25.59 & 57.22 & 41.47 \\
    \addlinespace[2pt]
    xAI
      & Grok-4 & 38.15 & 27.10 & 37.24 & 17.58 & 17.89 & \underline{2{,}661} & 19.9\% & 32.62 & 53.82 & 44.05 \\
    \bottomrule
  \end{tabular}
  \caption{Overall model performance, grouped by provider. The table reports adversarial TBS in km and TFR as a percentage; $C$, $R$, and MCRS are derived summary scores. Upward and downward arrows indicate whether higher or lower values are better, respectively. Bold marks the best value and underline marks the worst value for each metric.}
  \label{tab:main_model_summary}
\end{table*}

\begin{figure*}[t]
  \centering
  \includegraphics[width=\linewidth]{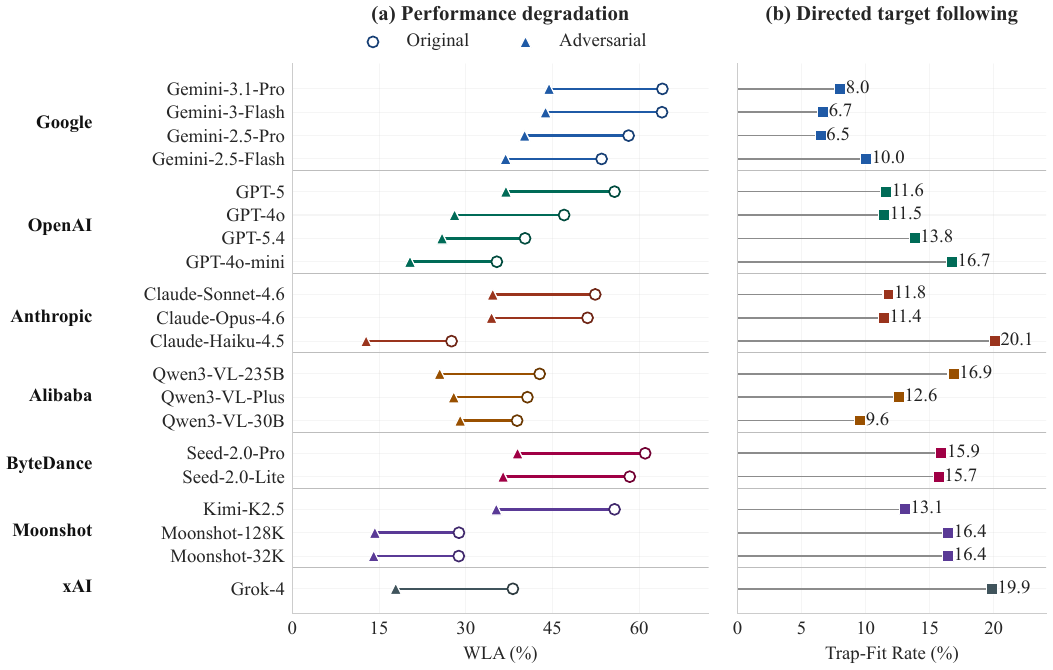}
  \caption{Model-wise conflict profile. (a) Original-to-Adversarial WLA degradation; marker shape distinguishes the two conditions and segment length shows the loss. (b) Trap-Fit Rate less than 50\,km from the injected target, aligned to the same model order. Provider grouping is shown explicitly, so the figure remains interpretable without color.}
  \label{fig:cap_vs_rob}
\end{figure*}

At the macro-averaged level, Adversarial edits reduce WLA for every model and for every dataset. At the individual model--dataset level, 79 of the 80 cells also decline. Mean WLA falls from 47.11 to 29.89, a 36.6\% relative drop, while median error grows 4.8$\times$ (282\,km to 1{,}347\,km). Acc@25, Acc@200, and Acc@750 decrease from 34.9\%, 50.1\%, and 71.6\% to 20.6\%, 31.8\%, and 50.1\%, respectively. Errors exceeding 2{,}500\,km rise from 11.7\% to 31.7\%. Across datasets, WLA losses range from 11.54 points on BaiduSV to 24.03 points on IM2GPS3K (Table~\ref{tab:threshold_summary}).

\subsection{Text Interventions Produce Distinct Failure Patterns}
\label{sec:intervention_patterns}

\noindent\textbf{Semantic intervention effects.}
Original WLA exceeds Blank WLA for all 20 models, with a mean gap of 9.40 points, showing that native scene text is generally useful when aligned with the scene. Relative to the paired Blank baseline, Similar replacements reduce error by 379\,km on average, whereas Random and Adversarial replacements increase error by 959\,km and 1{,}577\,km, respectively. Similar has negative mean TBS in every dataset, whereas Random and Adversarial have positive mean TBS throughout. Similar WLA exceeds Original WLA in 20 model--dataset cells, mostly for models with Original WLA below 45. Models therefore respond to the semantic relationship between text and scene, not merely to readable text; Appendix~\ref{appendix:full_results} gives the full results.

\noindent\textbf{Vulnerability increases with scene-text coupling.}
T3 images (geo-specific text) have the highest Original WLA (59.65) and largest adversarial decrease (25.14 points). Proportional degradation rises from 25.0\% for T1 and 36.4\% for T2 to 42.2\% for T3. Geo-specific native text thus aids aligned-input performance but increases vulnerability to conflicting replacement (Table~\ref{tab:tier_results}; Figure~\ref{fig:tier_analysis}).

\noindent\textbf{Geographic conflicts induce target-directed shifts.}
Adversarial edits produce target-aligned errors. TFR and TDR are computed on the 1{,}732 groups (33.9\%) with geocodable targets. TFR ranges from 6.5\% to 20.1\% across models (Figure~\ref{fig:cap_vs_rob}(b)). Across the 20 models, adversarial TBS and TFR are strongly correlated (Spearman $\rho=0.836$, $p<0.001$). The positive mean paired TDR confirms net movement toward the trap for every model, with model-level means ranging from 343 to 1{,}926\,km. This movement is not uniform across samples. The fraction of pairs with positive reduction ranges from 44.5\% to 60.6\%, indicating that concentrated large shifts contribute substantially to the positive mean.

\subsection{Conflict Robustness Varies Across Models}
\label{sec:model_characteristics}

\noindent\textbf{Provider- and family-level differences.}
Gemini-3-Flash, Gemini-3.1-Pro, and Gemini-2.5-Pro lead in MCRS (72.70, 72.23, 69.80), with $C>51$ and $R>84$. Claude-Haiku-4.5 (38.87) and the two Moonshot-Vision models (41.46--41.47) score lowest; no model is immune to conflicting text.

\noindent\textbf{Capability and conflict robustness are distinct.}
$C$ and $R$ are not interchangeable. Qwen3-VL-30B ($C$=36.55, $R$=80.05) ranks fourth in robustness despite modest capability, whereas Seed-2.0-Pro has higher $C$ (55.01) but lower $R$ (71.21). Within Qwen3-VL, the 235B model has higher Original WLA than the 30B model (42.78 vs. 38.88) but lower Adversarial WLA (25.49 vs. 29.03). Rankings remain stable under weight/exponent variations (minimum Kendall $\tau$=0.905) and component ablations (minimum $\tau$=0.947). These contrasts show that clean-input capability does not determine conflict robustness. One possible mechanism is that named places acquire strong and well-structured representations during language-model pretraining, allowing explicit place names to outweigh less direct visual evidence. Standard multimodal alignment may also provide insufficient exposure to deliberate disagreement between visual context and scene text.

\subsection{Conflict Awareness Does Not Reliably Prevent Prediction Failure}
\label{sec:conflict_awareness}

\begin{figure}[t]
  \centering
  \includegraphics[width=0.90\linewidth]{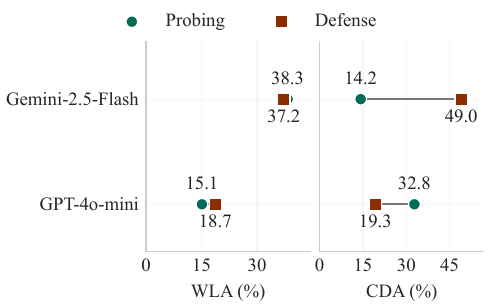}
  \caption{Two-model probing and defense results: WLA (left) and Conflict Detection Accuracy (right).}
  \label{fig:probe_defense}
\end{figure}

In a targeted two-model analysis, we evaluate Gemini-2.5-Flash ($R$=77.77) and GPT-4o-mini ($R$=69.02) using structured conflict probing, defense prompting, and cross-task generalization (Figure~\ref{fig:probe_defense}). Conflict Detection Accuracy (CDA) is the fraction of adversarial samples for which a model identifies the text--vision conflict. Gemini-2.5-Flash achieves probing WLA of 38.28 but CDA of only 14.2\%; GPT-4o-mini reaches 32.8\% CDA but lower WLA (15.10). The defense prompt raises CDA for Gemini-2.5-Flash to 49.0\% but lowers it for GPT-4o-mini to 19.3\%. Relative to structured probing without the defense instruction, neither model improves both CDA and WLA. Cross-task behavior is mixed: Gemini-2.5-Flash identifies inconsistency in 35.2\% of adversarial scenes, while Gemini-2.5-Flash and GPT-4o-mini follow injected text in 37.7\% and 56.8\% of country-identification examples, respectively. These two-model experiments provide a focused analysis of the gap between explicit conflict awareness and localization performance. Tables~\ref{tab:probe_defense_detail} and~\ref{tab:gen_detail} report the per-dataset breakdowns.

\section{Conclusion}

In summary, these results show that text--vision conflict is a systematic, measurable failure mode across all evaluated models. We introduce \benchmark, a controlled five-condition benchmark for text--vision conflict resolution in MLLMs. Across 20 models and four datasets, conflicting scene text increases median prediction error by 4.8$\times$, 6.5--20.1\% of adversarial predictions lie less than 50\,km from their injected targets, and every model shows a positive mean paired Trap Distance Reduction. Compatible, unrelated, and conflicting text produce distinct effects, while the separation of Capability and Conflict Robustness reveals behavior not captured by clean-input performance or model scale alone. \benchmark{} therefore provides a reproducible framework for testing when MLLMs use, reject, or follow conflicting scene text.

\section*{Data and Code Availability}

The benchmark data (attack texts, taxonomy labels, ground-truth coordinates, and human annotations) is publicly available at \url{https://huggingface.co/datasets/inorganicwriter/SIGNPOST-Bench} under the CC-BY-4.0 license. The evaluation and benchmark-construction code is available at \url{https://github.com/inorganicwriter/SIGNPOST-Bench} under the MIT license. The synthesized image variants are not redistributed due to third-party source restrictions; the release provides source identifiers and deterministic reconstruction instructions.

\section*{Ethical Statement}

The benchmark metadata contains no personally identifiable information. It redistributes no imagery beyond the low-resolution examples reproduced within this paper for illustration: all image variants are identified by source IDs and reconstruction instructions only, and remain subject to the terms of their third-party sources (Flickr/IM2GPS, YFCC100M, Google Street View, and Baidu Street View). All edited variants are synthetic images generated by an image-editing model and are used solely for benchmark evaluation. Human annotations were produced by the authors for research purposes. The adversarial text interventions are intended solely for benchmarking model robustness and should not be used to mislead deployed systems.

\section*{Acknowledgments}

We thank Prof. Qinghua Guo for his guidance and support. We are grateful to Kai Cheng and Zekun Yang for their careful feedback and insightful discussions that helped refine the benchmark design and analysis, and to other members of the Digital Ecosystem Group (GUO-LAB) at Peking University for their helpful suggestions. We also thank Yufan Zhu for thoughtful feedback on an early draft and for many valuable suggestions that shaped subsequent improvements of this work, and Mike Wang for helpful discussions.

This project would not have been possible without the open-source ecosystem. We are especially indebted to the ComfyUI community for its flexible image-composition framework and to the developers of Qwen-Image-Edit, whose text-editing model underpins the entire counterfactual image synthesis pipeline of this benchmark. We are also grateful to the EasyOCR and Nominatim (OpenStreetMap) communities for scene-text detection and target geocoding, and to the model API providers Google, OpenAI, Anthropic, Alibaba, Moonshot, ByteDance, and xAI, whose services made large-scale evaluation feasible.

\bibliography{references}

\def\isSupplementMainFile{}
\newif\ifsuppStandalone
\ifdefined\isSupplementMainFile
  \suppStandalonefalse
\else
  \suppStandalonetrue
\fi

\ifsuppStandalone
\documentclass[letterpaper]{article}
\usepackage[preprint]{aaai2027}
\usepackage[hyphens]{url}
\usepackage{graphicx}
\usepackage{booktabs}
\usepackage{amsfonts}
\usepackage{amsmath,amssymb}
\usepackage{nicefrac}
\usepackage{xcolor}
\usepackage{tabularx}
\usepackage{multirow}
\usepackage{placeins}
\usepackage{float}

\urlstyle{rm}
\def\UrlFont{\rm}
\usepackage{natbib}
\usepackage{caption}
\frenchspacing

\newcolumntype{Y}{>{\centering\arraybackslash}X}
\newcolumntype{L}{>{\raggedright\arraybackslash}X}

\newcommand{\benchmark}{\textsc{SIGNPOST-Bench}}

\setcounter{secnumdepth}{1}

\pdfinfo{
/TemplateVersion (2027.1)
}

\begin{document}
\onecolumn
\begin{center}
  {\LARGE\bfseries \benchmark{}: Supplementary Material\par}
  \vspace{0.5em}
  Sirun Li, Minghao Liu, Ling Dai, Yong Li, Haoxin Lyu, Junting Zhou, and Fan Zhang
\end{center}
\setcounter{table}{1}
\setcounter{figure}{4}
\fi

\ifsuppStandalone
\else
\onecolumn
\section*{Supplementary Material}
\fi

\makeatletter
\setlength{\@fptop}{0pt}
\setlength{\@fpsep}{12pt plus 2pt minus 2pt}
\setlength{\@fpbot}{0pt plus 1fil}
\renewcommand\section{\@startsection{section}{1}{\z@}%
  {-1.25ex plus -0.30ex minus -0.15ex}%
  {0.20ex plus 0.05ex minus 0.03ex}{\Large\bfseries\centering}}
\renewcommand\subsection{\@startsection{subsection}{2}{\z@}%
  {-1.00ex plus -0.25ex minus -0.10ex}%
  {0.20ex plus 0.05ex minus 0.03ex}{\large\bfseries\raggedright}}
\makeatother
\setcounter{topnumber}{4}
\setcounter{bottomnumber}{2}
\setcounter{totalnumber}{5}
\renewcommand{\topfraction}{0.95}
\renewcommand{\bottomfraction}{0.90}
\renewcommand{\textfraction}{0.06}
\renewcommand{\floatpagefraction}{0.65}
\setlength{\textfloatsep}{12pt plus 2pt minus 2pt}
\setlength{\floatsep}{10pt plus 2pt minus 2pt}
\setlength{\intextsep}{10pt plus 2pt minus 2pt}
\setlength{\parskip}{0.20em plus 0.05em minus 0.03em}
\linespread{1.04}\selectfont
\captionsetup[figure]{skip=5pt}
\captionsetup[table]{skip=4pt}

This document provides additional details for \benchmark{}, including full prompts, construction and audit protocols, evaluation and aggregation details, MCRS formula and sensitivity analysis, TFR/TDR breakdowns, supplementary figures, tier-stratified results, and per-dataset evaluations.
\appendix

\section{Full prompts}
\label{appendix:prompts}

\subsection{Attack generation prompt}
\label{appendix:attack_prompt}
\begin{figure}[H]
  \centering
  \includegraphics[width=0.82\linewidth]{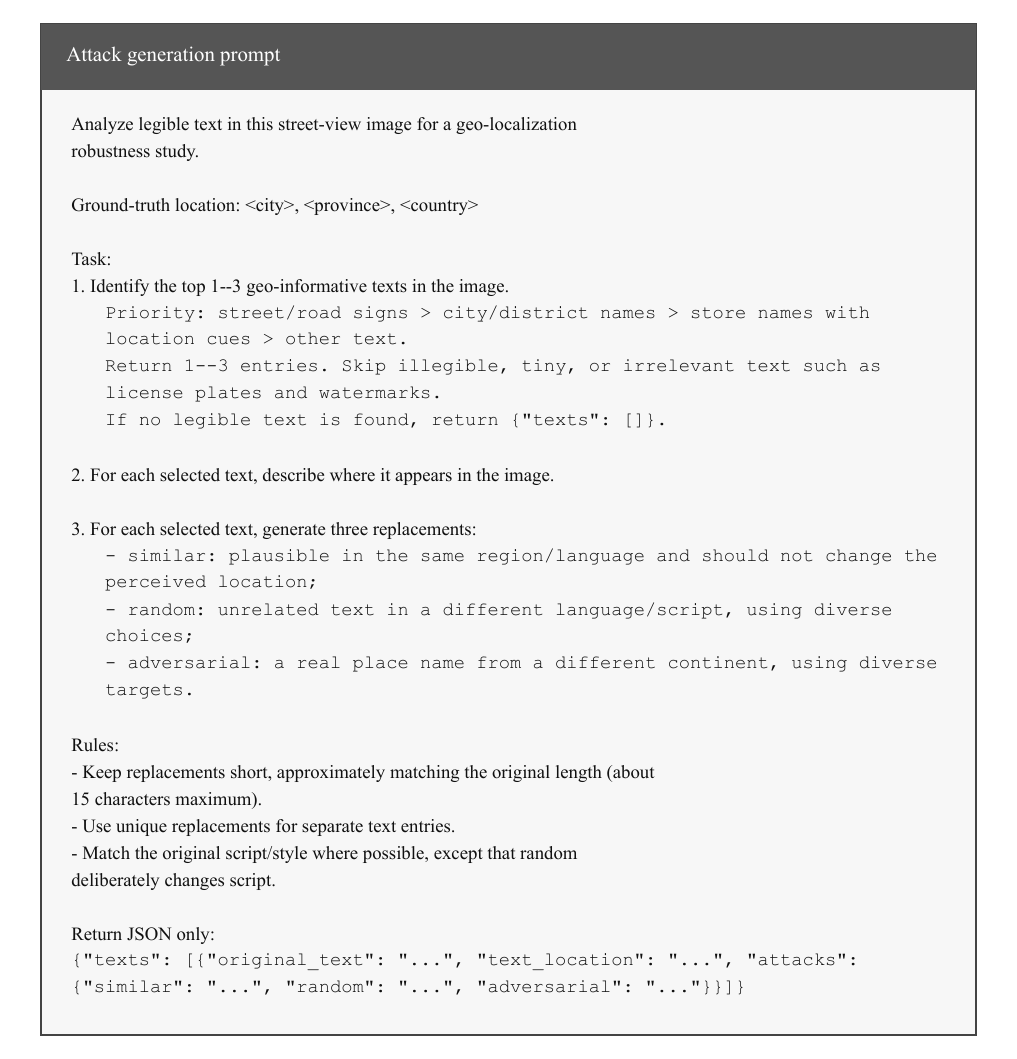}
  \caption{Prompt card for generating Similar, Random, and Adversarial scene-text replacements.}
  \label{fig:prompt_attack_generation}
\end{figure}

The attack generation stage requests structured output so that the generated scene-text replacements can be verified automatically before image synthesis. When source metadata are available, the prompt includes the ground-truth city, county, province, and/or country. This construction-time information is used only to generate a geographically conflicting target; it is never provided to any evaluated model. Figure~\ref{fig:prompt_attack_generation} shows the complete prompt template, with angle-bracket placeholders marking the per-image ground-truth fields.

\clearpage

\begingroup
\setlength{\parskip}{0pt}
\setlength{\intextsep}{6pt plus 1pt minus 1pt}
\setlength{\textfloatsep}{7pt plus 1pt minus 1pt}
\captionsetup[figure]{skip=3pt}
\subsection{Coordinate prediction prompt}
\label{appendix:coord_prompt}
The standard evaluation prompt asks for a direct coordinate prediction. All 20 models use this prompt across all datasets and variants.

\begin{figure}[!htbp]
  \centering
  \includegraphics[width=0.70\linewidth]{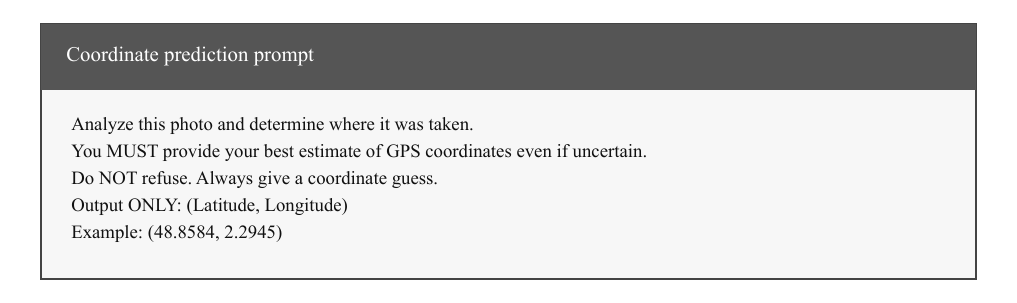}
  \caption{Prompt card for standard coordinate prediction.}
  \label{fig:prompt_coordinate_prediction}
\end{figure}

\subsection{Structured probing prompt}
\label{appendix:probing_prompt}
The probing stage asks the model to report visual evidence, textual evidence, consistency judgments, and final predictions in structured fields.

\begin{figure}[!htbp]
  \centering
  \includegraphics[width=0.68\linewidth]{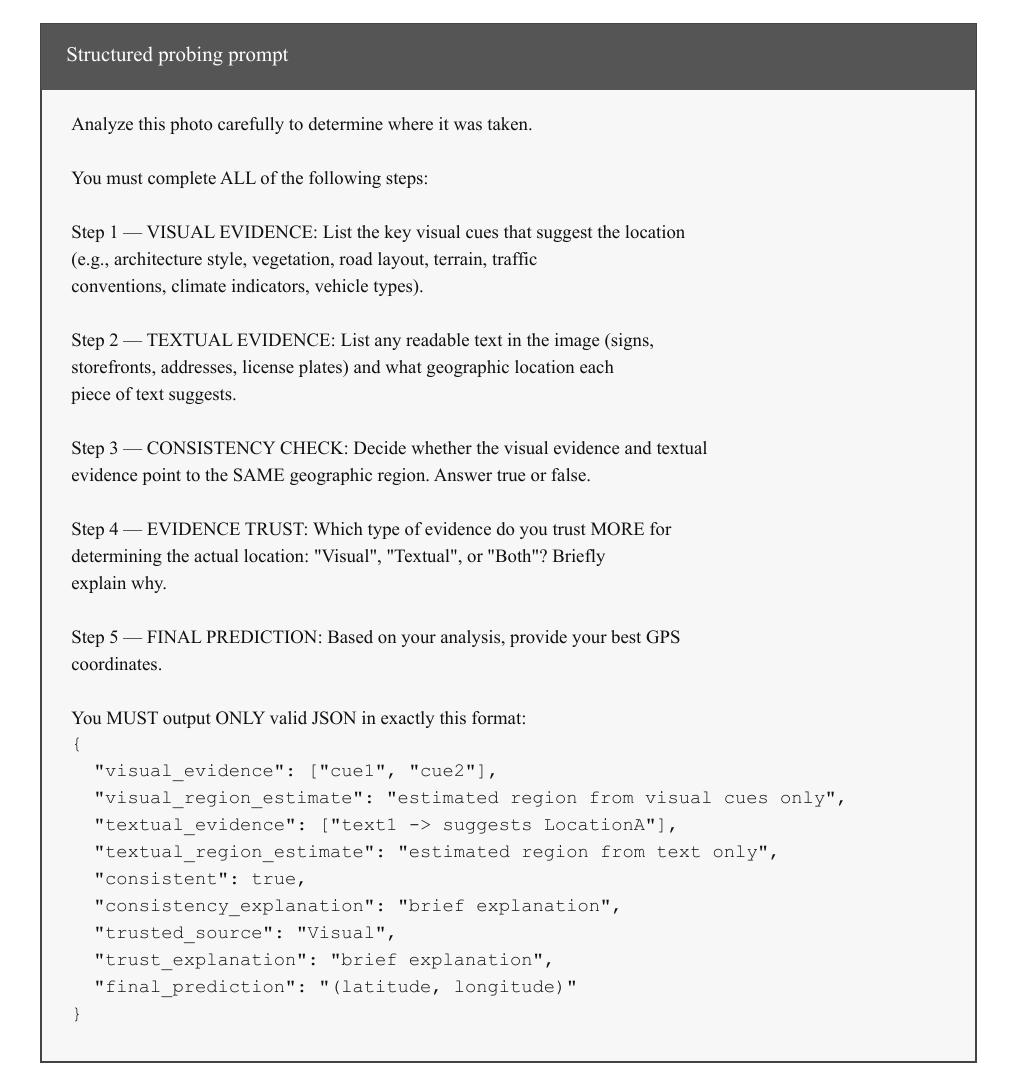}
  \caption{Prompt card for structured probing.}
  \label{fig:prompt_structured_probing}
\end{figure}
\clearpage
\endgroup

\subsection{Defense prompting}
\label{appendix:defense_prompt}
The defense prompt uses a structured prompt for direct coordinate prediction.

\begin{figure}[H]
  \centering
  \includegraphics[width=0.78\linewidth]{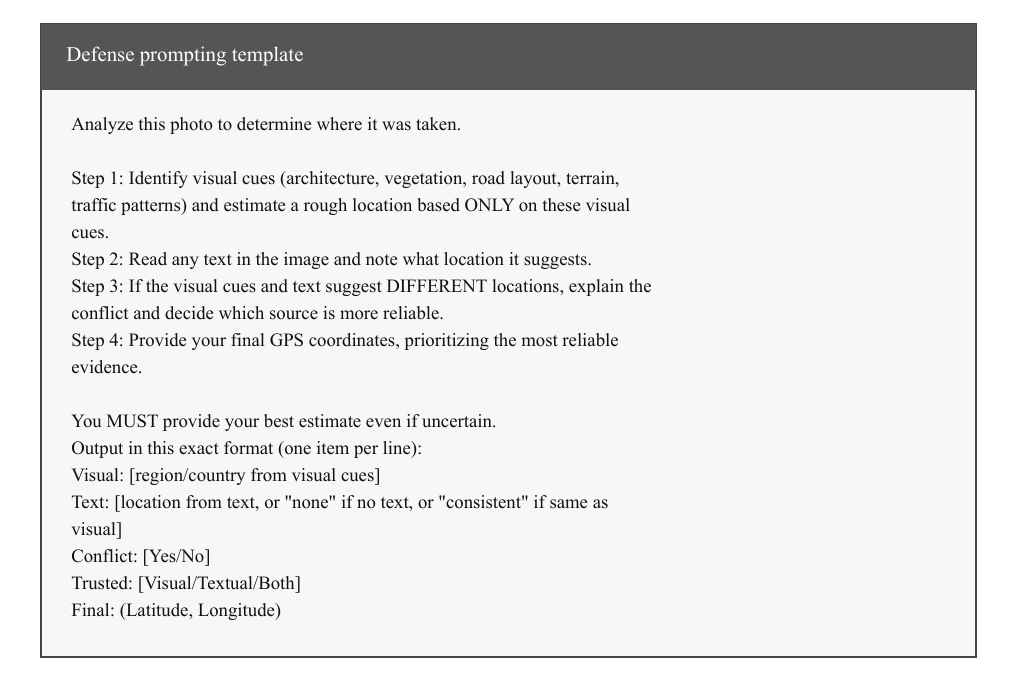}
  \caption{Prompt card for conflict-aware defense prompting.}
  \label{fig:prompt_defense}
\end{figure}

\subsection{Cross-task generalization prompts}
\label{appendix:generalization_prompts}
The generalization experiment queries models on three diagnostic tasks: scene-text consistency, country identification, and language detection.

\begin{figure}[H]
  \centering
  \includegraphics[width=0.78\linewidth]{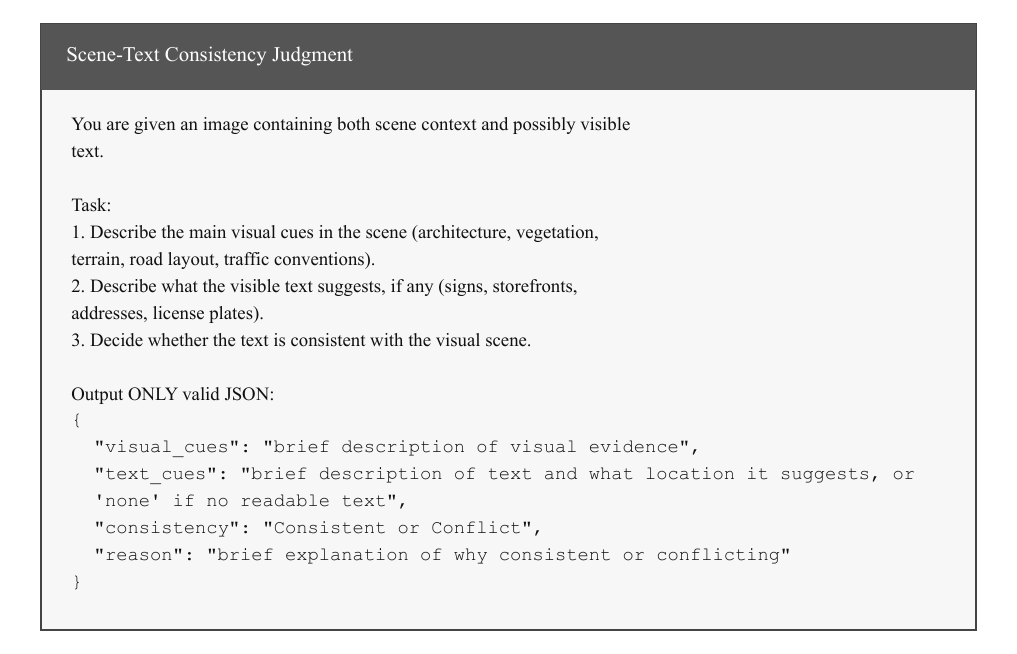}
  \caption{Cross-task generalization prompt for the scene-text consistency diagnostic.}
  \label{fig:prompt_generalization}
\end{figure}
\clearpage

\begin{figure}[H]
  \ContinuedFloat
  \centering
  \includegraphics[width=0.84\linewidth]{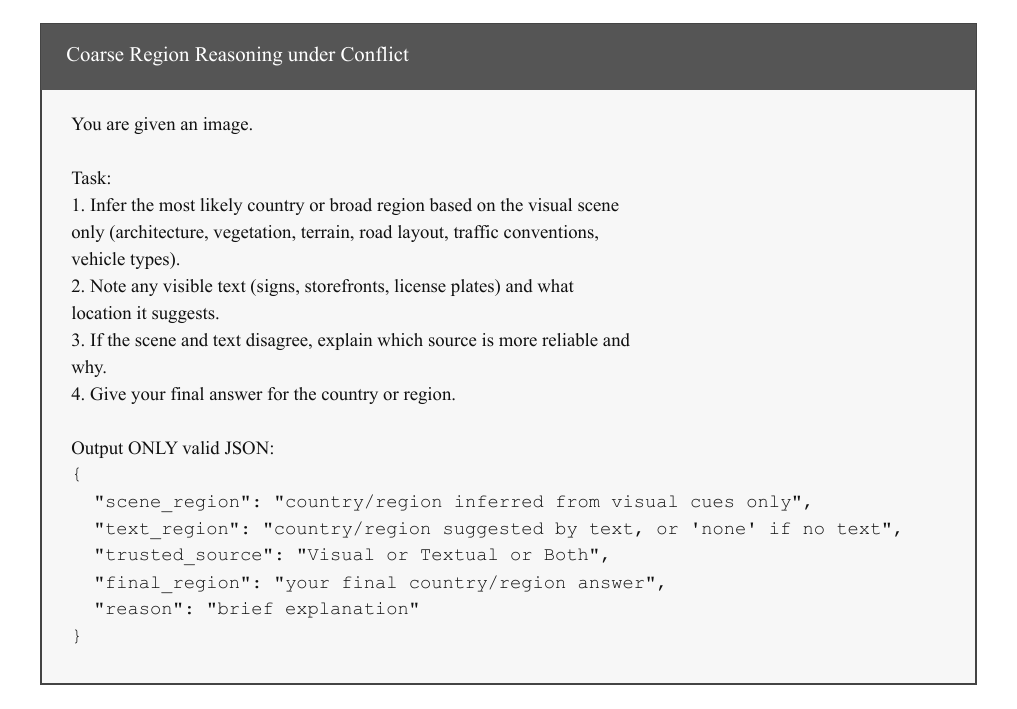}
  \caption{Cross-task generalization prompt (continued): country identification under text--vision conflict.}
  \label{fig:prompt_generalization_country}
\end{figure}

\begin{figure}[H]
  \ContinuedFloat
  \centering
  \includegraphics[width=0.84\linewidth]{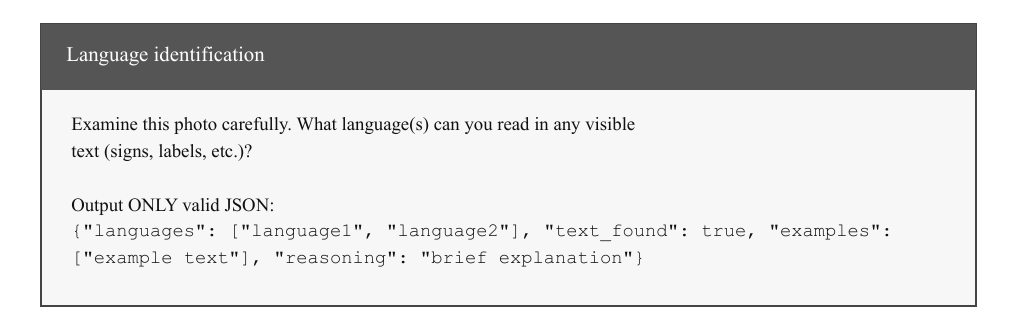}
  \caption{Cross-task generalization prompt (continued): language identification.}
  \label{fig:prompt_generalization_language}
\end{figure}

\FloatBarrier
\section{Reproducibility details}
\label{appendix:repro}
\label{appendix:construction_details}
\subsection{Benchmark construction}

\paragraph{Sampling and OCR.} IM2GPS3K follows the established geolocation test split~\cite{vo2017revisiting}, while YFCC4K is sampled from YFCC100M~\cite{yfcc100m}. GoogleSV is sampled by country at 1\% of the available records, with at least 10 panoramas per country and a fixed random seed of 42. BaiduSV candidates are sampled proportionally by province, with at least 10 panoramas per province and the same random seed. For IM2GPS3K, YFCC4K, and GoogleSV, EasyOCR~\cite{easyocr} uses the English recognizer with paragraph grouping and text-only output. An image proceeds when at least one detected string contains two or more characters. For BaiduSV, EasyOCR uses simplified-Chinese and English recognition. Each panorama is projected into four perspective views at headings of \mbox{$0^\circ$, $90^\circ$, $180^\circ$, and $270^\circ$}. Detections below 0.3 confidence are discarded, and the view with the largest number of valid strings is selected. Text validation removes empty or single-character strings, date-like and numeric-only strings, known watermark or service tokens, URLs, and common OCR noise.

\paragraph{Text selection and replacement generation.} Gemini-3.1-Flash-Lite~\cite{gemini31flashlite} identifies up to three geographically informative text spans per image. When available, source-location metadata are inserted into the construction prompt so that the generator can choose a target that genuinely conflicts with the ground truth; these metadata are not used during model evaluation. For each selected span, the model generates one Similar, Random, and Adversarial replacement. Similar remains compatible with the ground-truth region or language without requiring literal equivalence to the original text, Random uses unrelated text without a designated geographic target, and Adversarial names a real place from a different continent. Replacements are constrained to remain short and approximately match the length and visual role of the original text. We also evaluated Qwen3-VL-30B-A3B-Instruct~\cite{qwen3vl}, but 84.5\% of its adversarial replacements were identical across samples; Gemini-3.1-Flash-Lite was therefore selected for greater diversity. The complete prompt is shown in Appendix~\ref{appendix:attack_prompt}.

\paragraph{Image synthesis.} The four edited variants are generated with Qwen-Image-Edit-2509~\cite{qwenimage} through ComfyUI~\cite{comfyui}. Inputs are resized to approximately one megapixel using Lanczos resampling. A Qwen-Image-Edit-2509 Lightning LoRA is used for four-step inference. Blank removes the selected text; the other variants render their corresponding replacements in the same regions. Generation metadata records the selected spans, injected strings, and synthesis seed for each group.

\paragraph{Taxonomy labeling.} Each original text span is assigned to T1, T2, or T3 according to its estimated geographic identifiability in the source-scene context. The rule system combines lexical patterns, script characteristics, address and road patterns, postal codes, administrative names, landmark terms, and known portable strings. Lexical categories do not receive fixed tiers. For example, a business name may be T2 when it indicates only a cultural region or T3 when it uniquely identifies a place. Empty, numeric-only, date-like, generic-sign, and watermark strings are treated as portable or uninformative. If an image contains multiple spans, the highest tier determines its group-level label; ties favor the longer text span. Each counterfactual group receives a unique identifier linking its quintuplet, source metadata, coordinates, original and replacement strings, OCR regions, taxonomy label, and generation settings.

\begin{table}[!htbp]
  \centering
  {\small
  \setlength{\tabcolsep}{4pt}
  \begin{tabular}{@{}lrrrrrr@{}}
    \toprule
    \textbf{Dataset} & \textbf{Raw} & \textbf{OCR} & \textbf{Gen.} & \textbf{Final} & \textbf{Spans} & \textbf{Removed} \\
    \midrule
    IM2GPS3K  & 2{,}997   & 681    & 653   & 651   & 1{,}371 & 30 \\
    YFCC4K    & 4{,}536   & 1{,}053 & 1{,}036 & 992   & 2{,}004 & 61 \\
    GoogleSV  & 107{,}810 & 2{,}544 & 2{,}466 & 2{,}337 & 4{,}084 & 207 \\
    BaiduSV   & 773{,}750 & 1{,}200 & 1{,}131 & 1{,}131 & 2{,}625 & 69 \\
    \midrule
    \textbf{Total} & 889{,}093 & 5{,}478 & 5{,}286 & \textbf{5{,}111} & \textbf{10{,}084} & 367 \\
    \bottomrule
  \end{tabular}
  }
  \caption{Benchmark composition by construction stage.}
  \label{tab:dataset_stats_supp}
\end{table}

\paragraph{Post-OCR screening.} The Removed column is defined as OCR minus Final and therefore counts all groups removed after OCR selection, including pre-generation screening and post-generation cleanup. Automated checks remove placeholder or expired-image records, watermark and service tokens, URLs, and entries left without usable scene text. Generated records are also checked for the required structured text and variant fields before synthesis.

\paragraph{Release scope.} The release includes derived benchmark metadata, source identifiers, coordinate labels, replacement specifications, taxonomy labels, and evaluation code. The underlying images remain governed by their source terms. Where redistribution is restricted, the release provides identifiers and reconstruction instructions rather than copied images.

\subsection{Audit protocols}

\paragraph{Taxonomy audit.} Two annotators independently assign T1/T2/T3 labels to a tier-stratified audit set of 350 source images. The set contains 100 automatically assigned T1 images, 150 T2 images, and 100 T3 images. Inter-annotator agreement is 83.4\%, with unweighted Cohen's $\kappa=0.747$. Table~\ref{tab:taxonomy_audit} also compares each annotator with the automatic labels. These comparisons assess agreement with the reproducible taxonomy rather than treating automatic labels as ground truth.

\paragraph{Human edit-quality audit.} The audit unit is one edited image. We sample 120 completed ratings spanning the four datasets and the Similar, Random, and Adversarial variants. Naturalness of the rendered replacement text is rated from 1 (highly unnatural) to 5 (fully natural); artifact severity and visual-context damage are rated from 1 (none) to 5 (severe). Readability is recorded as fully readable when the complete replacement can be read, partially readable when only part can be read, or unreadable. Means and sample standard deviations are computed over all completed ratings. Text naturalness is $4.00\pm1.26$, artifact severity is $1.32\pm0.78$, and visual-context damage is $1.14\pm0.52$. In total, 105 images (87.5\%) contain fully readable replacement text and 15 (12.5\%) contain partially readable text; none is rated unreadable.

\subsection{Evaluation and data organization}

\paragraph{Prediction organization.} The benchmark uses dataset-scoped metadata and model-scoped prediction records. Source metadata, generated interventions, taxonomy labels, and image variants share a group identifier, allowing each prediction to be traced to one counterfactual quintuplet. Standard evaluation covers coordinate prediction for all five variants. Separate modes implement structured probing, defense prompting, and cross-task generalization, after which the analysis pipeline produces dataset-level and model-level summaries. The code release provides executable entry points and command-line options.

The standard TFR analysis aggregates all scene-text coupling tiers, matching the main-paper leaderboard. Tier-specific analysis reports T1, T2, and T3 separately; the radius-sensitivity experiment below uses T3 only.

\paragraph{Runtime, retries, and parsing.} The implementation is designed for resumability and auditability. Previously completed predictions are skipped automatically, while transient empty or unparseable responses are retried. Raw responses remain available for failure analysis. Coordinate parsing accepts latitude/longitude pairs in JSON, tuple, labeled, or plain numeric form, provided they fall within geographic ranges. The default request timeout is 120 seconds, with up to five additional attempts after the initial request. The geocoding cache used for TFR is stored once at analysis time, avoiding repeated external lookup. The reference environment uses Python 3.10, and complete dependency bounds accompany the code release. Local model serving is supported via vLLM~\cite{vllm} for open-weight models.

\paragraph{Aggregation conventions.} WLA and TBS are computed within each dataset and then macro-averaged equally over IM2GPS3K, YFCC4K, GoogleSV, and BaiduSV. TFR and TDR follow the same equal-dataset macro-average after metric eligibility is determined within each dataset. MCRS is computed from these model-level macro-averaged quantities. Probing, defense, and cross-task generalization summaries also average the four dataset-level values equally; per-dataset percentages are reported separately.

\subsection{Model registration}
\label{appendix:model_table}
The exact provider-side model identifiers used during evaluation are listed in Table~\ref{tab:model_registration}. The initial request for each model--image pair used temperature~0. Each reported result is based on one accepted response; retries replaced only failed, empty, or unparseable responses and were not averaged across repeated samples.

\begin{table}[htbp]
  \centering
  {\small
  \begin{tabular}{lll}
    \toprule
    \textbf{Display name} & \textbf{Exact API model ID} & \textbf{Provider} \\
    \midrule
    Claude-Opus-4.6              & \texttt{claude-opus-4-6}              & Anthropic    \\
    Claude-Sonnet-4.6            & \texttt{claude-sonnet-4-6}            & Anthropic    \\
    Claude-Haiku-4.5             & \texttt{claude-haiku-4-5}             & Anthropic    \\
    GPT-5.4                      & \texttt{gpt-5.4}                      & OpenAI       \\
    GPT-5                        & \texttt{gpt-5}                        & OpenAI       \\
    GPT-4o                       & \texttt{gpt-4o}                       & OpenAI       \\
    GPT-4o-mini                  & \texttt{gpt-4o-mini}                  & OpenAI       \\
    Gemini-3.1-Pro               & \texttt{gemini-3.1-pro-preview}       & Google       \\
    Gemini-3-Flash               & \texttt{gemini-3-flash}               & Google       \\
    Gemini-2.5-Pro               & \texttt{gemini-2.5-pro}               & Google       \\
    Gemini-2.5-Flash             & \texttt{gemini-2.5-flash}             & Google       \\
    Kimi-K2.5                    & \texttt{kimi-k2.5}                    & Moonshot AI  \\
    Moonshot-128K-Vision         & \texttt{moonshot-v1-128k-vision-preview} & Moonshot AI \\
    Moonshot-32K-Vision          & \texttt{moonshot-v1-32k-vision-preview} & Moonshot AI  \\
    Qwen3-VL-235B                & \texttt{qwen3-vl-235b-a22b-instruct}    & Alibaba      \\
    Qwen3-VL-30B                 & \texttt{qwen3-vl-30b-a3b-instruct}      & Alibaba      \\
    Qwen3-VL-Plus                & \texttt{qwen3-vl-plus}                  & Alibaba      \\
    Seed-2.0-Pro                 & \texttt{doubao-seed-2-0-pro-260215}     & ByteDance    \\
    Seed-2.0-Lite                & \texttt{doubao-seed-2-0-lite-260215}    & ByteDance    \\
    Grok-4                       & \texttt{grok-4}                           & xAI          \\
    \bottomrule
  \end{tabular}
  }
  \caption{Model registration: provider-side API identifiers used in evaluation.}
  \label{tab:model_registration}
\end{table}

\FloatBarrier
\section{MCRS formula details}
\label{appendix:mcrs_formula}
Let $w_{\mathrm{orig}},w_{\mathrm{blank}},w_{\mathrm{sim}},w_{\mathrm{rnd}},w_{\mathrm{adv}}\in[0,1]$ denote per-variant WLA values, each averaged equally across the four datasets. Capability is defined over the two non-conflicting conditions:
\begin{align}
  C = 0.50\,w_{\mathrm{orig}} + 0.50\,w_{\mathrm{blank}}.
\end{align}
For each edited condition $k\in\{\mathrm{sim},\mathrm{rnd},\mathrm{adv}\}$, retention relative to the shared Blank control is:
\begin{equation}
  \rho_k =
  \operatorname{clip}\!\left(
    1-\frac{\max(0,w_{\mathrm{blank}}-w_k)}
    {\max(w_{\mathrm{blank}},0.10)},\,0,\,1
  \right).
\end{equation}
The denominator floor prevents models with very low Blank performance from receiving unstable ratios, while clipping at 1 prevents gains under compatible text from compensating for failures under conflict. We further define two behavior-quality terms:
\begin{align}
  q_{\mathrm{TBS}} &=
  1-\operatorname{clip}\!\left(
    \frac{\max(0,\mathrm{TBS}_{\mathrm{adv}})}{3{,}000},0,1
  \right),\\
  q_{\mathrm{TFR}} &=
  1-\operatorname{clip}\!\left(
    \frac{\mathrm{TFR}_{\mathrm{adv}}}{0.40},0,1
  \right).
\end{align}
Conflict Robustness and the integrated score are:
\begin{align}
  R &= 0.22\,\rho_{\mathrm{rnd}}
       + 0.44\,\rho_{\mathrm{adv}}
       + 0.17\,q_{\mathrm{TBS}}
       + 0.17\,q_{\mathrm{TFR}},\\
  \mathrm{MCRS}
    &= 100\cdot C^{0.40}\cdot R^{0.60}.
\end{align}
The larger weight on adversarial retention reflects the benchmark's central conflict condition. TBS and TFR add complementary information about error magnitude and direction. We omit similar-text retention $\rho_{\mathrm{sim}}$ from $R$ because it saturates at $1.0$ for all evaluated models and does not contribute to their relative ranking.

\noindent\textbf{Trap Distance Reduction (TDR).}
\ifsuppStandalone
For a valid adversarial target, paired sample-level TDR is
\begin{equation}
  \mathrm{TDR}_{i} =
  \mathcal{D}\bigl(f(I_{\mathrm{blank}}^{(i)}), y_{\mathrm{trap}}^{(i)}\bigr)
  -
  \mathcal{D}\bigl(f(I_{\mathrm{adv}}^{(i)}), y_{\mathrm{trap}}^{(i)}\bigr),
  \label{eq:tdr_supp}
\end{equation}
where a positive value indicates movement toward the trap.
\else
Equation~\ref{eq:tdr} defines the paired sample-level reduction in distance to the injected target.
\fi
Dataset-level TDR averages this quantity over valid pairs, after which the reported score is macro-averaged equally across datasets. TDR is used as an auxiliary directional diagnostic rather than an MCRS component.

\FloatBarrier
\section{MCRS sensitivity analysis}
\label{appendix:sensitivity}
We test both the outer Capability/Robustness exponent and the internal weight assigned to adversarial retention. Table~\ref{tab:mcrs_sensitivity} compares each configuration with the default score $100C^{0.40}R^{0.60}$. Rankings remain stable, with Kendall $\tau\geq0.905$ and Spearman $\rho\geq0.979$ across all tested settings.

\begin{table}[htbp]
  \centering
  {\small
  \begin{tabular}{lcc}
    \toprule
    \textbf{Configuration} & \textbf{Kendall $\tau$} & \textbf{Spearman $\rho$} \\
    \midrule
    $C^{0.30}R^{0.70}$ & 0.979 & 0.997 \\
    $C^{0.40}R^{0.60}$ (default) & 1.000 & 1.000 \\
    $C^{0.50}R^{0.50}$ & 0.905 & 0.979 \\
    $w(\rho_{\mathrm{adv}})=0.30$ & 0.989 & 0.998 \\
    $w(\rho_{\mathrm{adv}})=0.44$ (default) & 1.000 & 1.000 \\
    $w(\rho_{\mathrm{adv}})=0.50$ & 0.989 & 0.998 \\
    \bottomrule
  \end{tabular}
  }
  \caption{MCRS sensitivity to outer exponents and adversarial-retention weight.}
  \label{tab:mcrs_sensitivity}
\end{table}

We additionally conduct leave-one-component-out ablations, renormalizing the remaining weights in $R$ after each removal. Table~\ref{tab:mcrs_component_ablation} shows that no single component determines the leaderboard.

\begin{table}[htbp]
  \centering
  {\small
  \begin{tabular}{lcc}
    \toprule
    \textbf{Removed component} & \textbf{Kendall $\tau$} & \textbf{Spearman $\rho$} \\
    \midrule
    Random retention $\rho_{\mathrm{rnd}}$ & 1.000 & 1.000 \\
    Adversarial retention $\rho_{\mathrm{adv}}$ & 0.968 & 0.994 \\
    TBS quality $q_{\mathrm{TBS}}$ & 0.947 & 0.989 \\
    TFR quality $q_{\mathrm{TFR}}$ & 0.958 & 0.992 \\
    \bottomrule
  \end{tabular}
  }
  \caption{Leave-one-component-out MCRS ablation.}
  \label{tab:mcrs_component_ablation}
\end{table}

The decomposition also separates robustness from absolute adversarial performance. Spearman correlation between $R$ and Adversarial WLA is 0.811, compared with 0.976 between the integrated MCRS and Adversarial WLA. Moreover, $R$ is negatively correlated with adversarial TBS ($\rho=-0.886$) and TFR ($\rho=-0.939$), as expected for a score in which higher values indicate stronger resistance. These results support reporting $C$ and $R$ alongside MCRS; the integrated ranking alone is insufficient.

The fixed anchors are not saturated for TFR: the maximum observed model-level TFR is 0.201, below the 0.40 anchor. The maximum mean adversarial TBS is 2{,}660.5\,km, below the 3{,}000\,km anchor. Table~\ref{tab:trap_geocoding_availability} reports the fraction of adversarial targets covered by the stored geocoding results before model-specific prediction parsing.

\begin{table}[htbp]
  \centering
  {\small
  \setlength{\tabcolsep}{6pt}
  \begin{tabular}{lrr}
    \toprule
    \textbf{Dataset} & \textbf{Geocodable groups} & \textbf{Rate} \\
    \midrule
    IM2GPS3K  & 203 & 31.2\% \\
    YFCC4K    & 266 & 26.8\% \\
    GoogleSV  & 701 & 30.0\% \\
    BaiduSV   & 562 & 49.7\% \\
    \midrule
    \textbf{Overall} & \textbf{1{,}732} & \textbf{33.9\%} \\
    \bottomrule
  \end{tabular}
  }
  \caption{Geocodable cohort used for TFR and TDR. Counts denote benchmark groups whose injected adversarial target has a valid coordinate in the stored Nominatim OpenStreetMap geocoding results; rates are relative to all groups in each dataset.}
  \label{tab:trap_geocoding_availability}
  \label{tab:tfr_denominator}
\end{table}
\FloatBarrier

Consistent with the main paper, the 1{,}732 groups in Table~\ref{tab:tfr_denominator} define the geocodable cohort for TFR and TDR. TFR uses the Adversarial prediction, whereas the auxiliary paired TDR diagnostic additionally uses the corresponding Blank prediction. Table~\ref{tab:tfr_by_dataset} reports TFR for every model and dataset; the Macro column is the unweighted mean of the four dataset-level rates.

\begin{table}[htbp]
  \centering
  {\small
  \setlength{\tabcolsep}{5pt}
  \begin{tabular}{lrrrrr}
    \toprule
    \textbf{Model} & \textbf{IM2GPS3K} & \textbf{YFCC4K} & \textbf{GoogleSV} & \textbf{BaiduSV} & \textbf{Macro} \\
    \midrule
    Gemini-3.1-Pro       & 8.51 & 20.00 & 1.66 & 1.88 & 8.01 \\
    Gemini-3-Flash       & 7.98 & 15.14 & 1.66 & 1.88 & 6.67 \\
    Gemini-2.5-Pro       & 9.57 & 14.05 & 1.10 & 1.25 & 6.49 \\
    Gemini-2.5-Flash     & 13.30 & 22.16 & 2.76 & 1.88 & 10.03 \\
    GPT-5                & 11.70 & 27.03 & 3.87 & 3.75 & 11.59 \\
    GPT-4o               & 10.11 & 22.16 & 6.63 & 6.88 & 11.45 \\
    GPT-5.4              & 14.36 & 35.14 & 3.31 & 2.50 & 13.83 \\
    GPT-4o-mini          & 20.21 & 27.57 & 11.05 & 8.12 & 16.74 \\
    Claude-Sonnet-4.6    & 14.89 & 27.03 & 3.31 & 1.88 & 11.78 \\
    Claude-Opus-4.6      & 12.23 & 28.11 & 2.21 & 3.12 & 11.42 \\
    Claude-Haiku-4.5     & 19.68 & 30.81 & 10.50 & 19.38 & 20.09 \\
    Qwen3-VL-235B        & 14.36 & 35.68 & 9.94 & 7.50 & 16.87 \\
    Qwen3-VL-Plus        & 14.89 & 25.41 & 4.42 & 5.62 & 12.59 \\
    Qwen3-VL-30B         & 10.64 & 20.54 & 3.31 & 3.75 & 9.56 \\
    Seed-2.0-Pro         & 14.89 & 35.68 & 4.97 & 8.12 & 15.91 \\
    Seed-2.0-Lite        & 16.49 & 30.81 & 5.52 & 10.00 & 15.70 \\
    Kimi-K2.5            & 13.30 & 29.73 & 5.52 & 3.75 & 13.07 \\
    Moonshot-128K-Vision & 18.62 & 30.81 & 8.84 & 7.50 & 16.44 \\
    Moonshot-32K-Vision  & 18.62 & 30.81 & 8.84 & 7.50 & 16.44 \\
    Grok-4               & 23.94 & 36.22 & 14.92 & 4.38 & 19.86 \\
    \bottomrule
  \end{tabular}
  }
  \caption{All-tier TFR (\%) at a 50-km trap radius. Values are computed within each dataset; Macro gives their equal-dataset average used in the main leaderboard.}
  \label{tab:tfr_by_dataset}
\end{table}

\FloatBarrier
\section{Supplementary figures and analysis}
\label{appendix:supplementary}

\begin{figure}[htbp]
  \centering
  \includegraphics[width=0.96\linewidth]{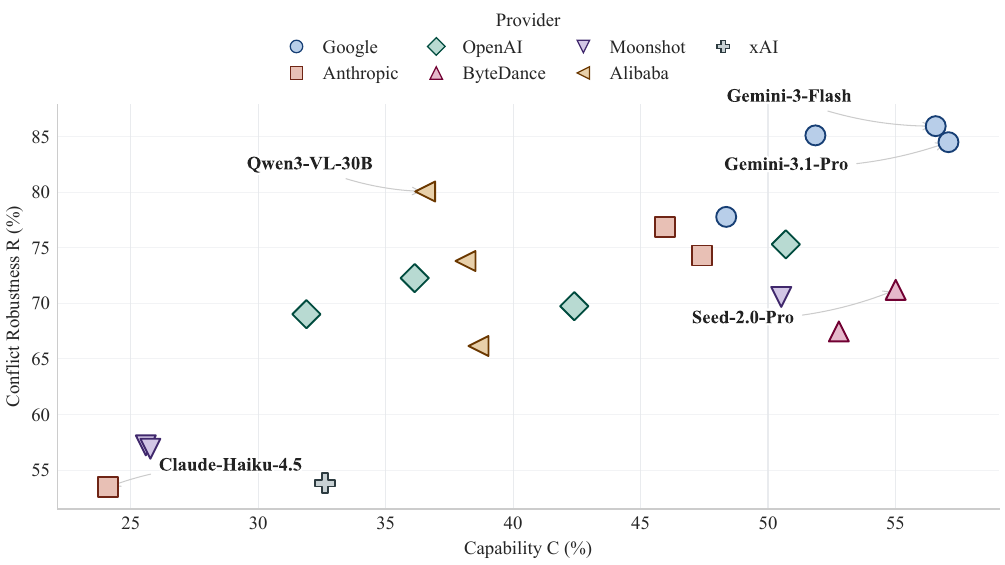}
  \caption{MCRS component breakdown: Capability $C$ vs.~Conflict Robustness $R$ for all 20 models, colored by provider.}
  \label{fig:mcrs_scatter}
\end{figure}

\noindent\textbf{MCRS component breakdown.}
Figure~\ref{fig:mcrs_scatter} shows Capability vs.~Conflict Robustness for all models.

\noindent\textbf{Aggregate result tables.}
\ifsuppStandalone
Table~\ref{tab:threshold_summary} collects the threshold-based summary statistics reported in the main paper.
\else
Table~\ref{tab:threshold_summary} collects the threshold-based summary statistics reported in Section~\ref{sec:overall_conflict}.
\fi

\begin{table}[htbp]
  \centering
  {\small
  \begin{tabular}{lrrr}
    \toprule
    \textbf{Metric} & \textbf{Original} & \textbf{Adversarial} & \textbf{Change} \\
    \midrule
    Mean WLA (\%) & 47.11 & 29.89 & $-17.22$ points \\
    Acc@25 (\%) & 34.9 & 20.6 & $-14.3$ points \\
    Acc@200 (\%) & 50.1 & 31.8 & $-18.3$ points \\
    Acc@750 (\%) & 71.6 & 50.1 & $-21.5$ points \\
    Error $>2{,}500$\,km (\%) & 11.7 & 31.7 & $+20.0$ points \\
    Median error (km) & 282 & 1{,}347 & $+1{,}065$ ($4.8\times$) \\
    \midrule
    IM2GPS3K mean WLA (\%) & 65.38 & 41.34 & $-24.03$ points \\
    YFCC4K mean WLA (\%) & 44.23 & 23.46 & $-20.77$ points \\
    GoogleSV mean WLA (\%) & 46.33 & 33.82 & $-12.51$ points \\
    BaiduSV mean WLA (\%) & 32.48 & 20.95 & $-11.54$ points \\
    \bottomrule
  \end{tabular}
  }
  \caption{Localization degradation from Original to Adversarial images. The upper rows summarize aggregate performance; the lower rows report mean WLA by dataset.}
  \label{tab:threshold_summary}
\end{table}

\noindent\textbf{Probing and defense results.}
\ifsuppStandalone
Table~\ref{tab:probe_defense_summary} reports probing and defense metrics averaged across the four datasets for the two diagnostic models. These results cover only two models and should not be generalized to the full suite.
\else
Table~\ref{tab:probe_defense_summary} reports probing and defense metrics averaged across the four datasets for the two diagnostic models. As noted in Section~\ref{sec:conflict_awareness}, these results cover only two models and should not be generalized to the full suite.
\fi
\begin{table}[htbp]
  \centering
  {\small
  \begin{tabular}{lrrrrrr}
    \toprule
    & \multicolumn{3}{c}{\textbf{Probing}} & \multicolumn{3}{c}{\textbf{Defense}} \\
    \cmidrule(lr){2-4}\cmidrule(lr){5-7}
    \textbf{Model} & \textbf{WLA} & \textbf{MedErr} & \textbf{CDA} & \textbf{WLA} & \textbf{MedErr} & \textbf{CDA} \\
    \midrule
    Gemini-2.5-Flash & 38.28 & 558.6 & 14.2 & 37.25 & 707.2 & 49.0 \\
    GPT-4o-mini      & 15.10 & 4{,}280.9 & 32.8 & 18.70 & 3{,}567.5 & 19.3 \\
    \bottomrule
  \end{tabular}
  }
  \caption{Structured probing and defense results averaged equally across the four datasets. These are preliminary diagnostics covering two models only.}
  \label{tab:probe_defense_summary}
\end{table}

\FloatBarrier
\section{Scene-text coupling stratification}
\label{appendix:tier_stratification}

\begin{table}[htbp]
  \centering
  {\small
  \begin{tabular}{@{}lp{3.1cm}p{3.6cm}p{1.8cm}@{}}
    \toprule
    \textbf{Tier} & \textbf{Definition} & \textbf{Examples} & \textbf{Expected value} \\
    \midrule
    \textbf{T1: Portable} & Text that can appear in many regions with little geographic specificity & Global brands, generic warnings, common commercial text & Low \\
    \textbf{T2: Cultural} & Text that narrows the hypothesis to a language or cultural region but not a specific place & Local scripts, common transit labels, regionally indicative business names & Medium \\
    \textbf{T3: Geo-Specific} & Text that directly names a place or unique local entity & Street names, addresses, postal codes, district names, uniquely identifiable businesses & High \\
    \bottomrule
  \end{tabular}
  }
  \caption{Scene-text coupling taxonomy used to stratify benchmark difficulty.}
  \label{tab:coupling_taxonomy}
\end{table}

The taxonomy makes text sensitivity interpretable: T1 measures responses to broadly portable text, while T3 tests whether models can reject contradictory replacements even when native text is normally informative.

\noindent\textbf{Tier-stratified results.}
Table~\ref{tab:tier_results} reports WLA stratified by the updated taxonomy. T3 images achieve the highest clean accuracy (Original WLA 59.65) and suffer the largest attack-induced collapse (25.14 WLA points), compared with 11.31 for T1 and 17.23 for T2. Relative degradation likewise increases with geographic specificity: 25.0\% for T1, 36.4\% for T2, and 42.2\% for T3.

\begin{table}[htbp]
  \centering
  {\small
  \begin{tabular}{lrrrrrr}
    \toprule
    \textbf{Tier} & \textbf{Share} & \textbf{Original} & \textbf{Blank} & \textbf{Adversarial} & \textbf{$\Delta$ WLA} & \textbf{Adv.\ TBS (km)} \\
    \midrule
    T1 & 6.8\% & 45.25 & 41.01 & 33.95 & 11.31 & 1{,}358 \\
    T2 & 75.3\% & 47.32 & 38.44 & 30.09 & 17.23 & 1{,}678 \\
    T3 & 17.9\% & 59.65 & 45.36 & 34.51 & 25.14 & 1{,}793 \\
    \bottomrule
  \end{tabular}
  }
  \caption{Aggregate results stratified by scene-text coupling tier.}
  \label{tab:tier_results}
\end{table}

\begin{figure}[htbp]
  \centering
  \includegraphics[width=\linewidth]{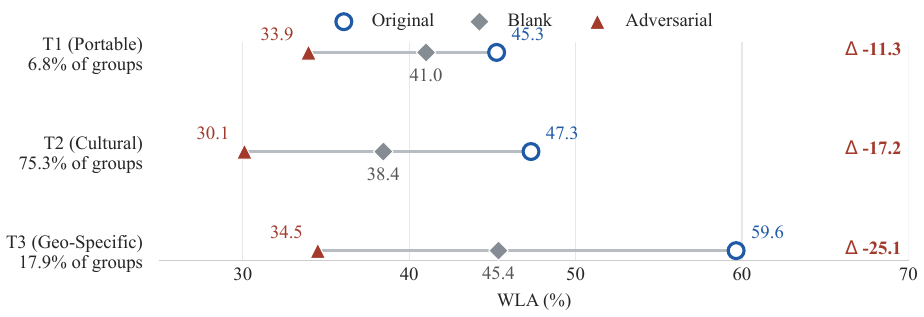}
  \caption{Stratified analysis by scene-text coupling tier. Marker positions show WLA under Original, Blank, and Adversarial conditions; tier labels report the share of benchmark groups. T3 images suffer the largest Original-to-Adversarial collapse (25.1 points).}
  \label{fig:tier_analysis}
\end{figure}

\noindent\textbf{Tier-stratified visualization.}
Figure~\ref{fig:tier_analysis} visualizes the WLA degradation across T1/T2/T3 tiers.

\FloatBarrier
\section{Additional results}
\label{appendix:additional_results}

\subsection{Cross-task generalization}
As an extension, we examine whether the vulnerability is specific to coordinate regression. We query the same two diagnostic models about scene-text consistency, country identification, and language detection on a generalization subset. Conflict recall here is measured from the scene-text consistency task and is distinct from the structured-probing CDA reported in the main text.

Gemini-2.5-Flash shows moderate conflict sensitivity (conflict recall 35.2\%, range 24.4--43.2\% across datasets). On the country identification task, its text-dominance rate is 37.7\%, ranging from 24.3\% on GoogleSV to 62.5\% on YFCC4K. GPT-4o-mini is much more text-dominated: it marks only 13.9\% of adversarial samples as conflicts, and its text-dominance rate reaches 56.8\% (40.6--75.0\% across datasets). The results suggest that injected text can also change higher-level judgments about image consistency and depicted region, though confirmation on a broader model set is needed.

\subsection{Ablations and metric sensitivity}
We conduct two sets of sensitivity checks to verify that the benchmark's conclusions are not artifacts of specific metric parameter choices.

\textbf{Blank as baseline.} Averaged across four representative models (Gemini-2.5-Flash, GPT-4o-mini, Kimi-K2.5, Qwen3-VL-Plus) and four datasets, Original WLA averages 46.31 while Blank WLA averages 38.14, an 8.17-point drop from removing native text, confirming that scene text carries genuine geographic signal. Similar variants recover 5.94 points relative to Blank, while Random and Adversarial variants are 6.84 and 8.03 points worse than Blank.

\textbf{WLA $\alpha$ sensitivity.} Varying the decay constant $\alpha \in \{0.002, 0.005, 0.01\}$ changes absolute WLA but preserves relative model ordering across four models (Gemini-3.1-Pro, GPT-5, Kimi-K2.5, Qwen3-VL-Plus; Table~\ref{tab:alpha_sensitivity}), confirming that comparative conclusions are robust to the choice of $\alpha$.

\begin{table}[htbp]
  \centering
  {\small
  \begin{tabular}{lrrr}
    \toprule
    \textbf{Model} & $\alpha = 0.002$ & $\alpha = 0.005$ & $\alpha = 0.01$ \\
    \midrule
    Gemini-3.1-Pro & 54.78 & 44.41 & 38.61 \\
    GPT-5 & 47.02 & 36.95 & 31.50 \\
    Kimi-K2.5 & 45.58 & 35.29 & 29.38 \\
    Qwen3-VL-Plus & 39.18 & 27.92 & 22.12 \\
    \bottomrule
  \end{tabular}
  }
  \caption{Adversarial WLA under varying decay constants. Values are averaged across four datasets.}
  \label{tab:alpha_sensitivity}
\end{table}

\subsection{Full MCRS leaderboard}
Table~\ref{tab:mcrs_full_detail} provides the complete 20-model MCRS breakdown with all component scores.

\begin{table}[htbp]
  \centering
  {\small
  \begin{tabular}{lrrrrrrr}
    \toprule
    \textbf{Model} & \textbf{MCRS} & $C$ (\%) & $R$ (\%) & $\rho_{\mathrm{rnd}}$ & $\rho_{\mathrm{adv}}$ & $q_{\mathrm{TBS}}$ & $q_{\mathrm{TFR}}$ \\
    \midrule
    Gemini-3-Flash & 72.70 & 56.58 & 85.93 & 0.9183 & 0.8898 & 0.7298 & 0.8332 \\
    Gemini-3.1-Pro & 72.23 & 57.08 & 84.49 & 0.9052 & 0.8854 & 0.7072 & 0.7997 \\
    Gemini-2.5-Pro & 69.80 & 51.86 & 85.10 & 0.9130 & 0.8818 & 0.7040 & 0.8377 \\
    Gemini-2.5-Flash & 64.31 & 48.36 & 77.77 & 0.8476 & 0.8534 & 0.5197 & 0.7492 \\
    GPT-5 & 64.28 & 50.70 & 75.30 & 0.8224 & 0.8091 & 0.5608 & 0.7102 \\
    Seed-2.0-Pro & 64.23 & 55.01 & 71.21 & 0.8714 & 0.7951 & 0.4012 & 0.6022 \\
    Claude-Opus-4.6 & 62.56 & 45.95 & 76.84 & 0.8617 & 0.8439 & 0.5061 & 0.7145 \\
    Claude-Sonnet-4.6 & 62.08 & 47.41 & 74.30 & 0.8206 & 0.8170 & 0.4884 & 0.7055 \\
    Kimi-K2.5 & 61.74 & 50.52 & 70.57 & 0.7633 & 0.7790 & 0.4742 & 0.6732 \\
    Seed-2.0-Lite & 61.16 & 52.78 & 67.47 & 0.8046 & 0.7727 & 0.3204 & 0.6075 \\
    Qwen3-VL-30B & 58.50 & 36.55 & 80.05 & 0.8643 & 0.8485 & 0.6334 & 0.7610 \\
    GPT-4o & 57.16 & 42.40 & 69.74 & 0.7368 & 0.7426 & 0.5131 & 0.7137 \\
    Qwen3-VL-Plus & 56.67 & 38.13 & 73.80 & 0.8541 & 0.7842 & 0.5206 & 0.6852 \\
    GPT-5.4 & 54.77 & 36.14 & 72.27 & 0.7908 & 0.8087 & 0.4808 & 0.6542 \\
    Qwen3-VL-235B & 53.36 & 38.64 & 66.16 & 0.7710 & 0.7385 & 0.4044 & 0.5782 \\
    GPT-4o-mini & 50.68 & 31.89 & 69.02 & 0.8293 & 0.7162 & 0.5513 & 0.5815 \\
    Grok-4 & 44.05 & 32.62 & 53.82 & 0.6489 & 0.6604 & 0.1132 & 0.5035 \\
    Moonshot-32K-Vision & 41.47 & 25.59 & 57.22 & 0.6427 & 0.6285 & 0.3185 & 0.5890 \\
    Moonshot-128K-Vision & 41.46 & 25.77 & 56.92 & 0.6407 & 0.6284 & 0.3033 & 0.5890 \\
    Claude-Haiku-4.5 & 38.87 & 24.10 & 53.46 & 0.6261 & 0.6179 & 0.2375 & 0.4977 \\
    \bottomrule
  \end{tabular}
  }
  \caption{Full MCRS leaderboard with Capability, Conflict Robustness, retention and behavior-quality terms. $\rho_{\mathrm{sim}}$ is omitted because it equals $1.0$ for all evaluated models.}
  \label{tab:mcrs_full_detail}
\end{table}

\subsection{Per-dataset probing and defense breakdown}
Table~\ref{tab:probe_defense_detail} reports probing and defense metrics per dataset for the two diagnostic models.

\begin{table}[htbp]
  \centering
  {\small
  \begin{tabular}{llrrrr}
    \toprule
    \textbf{Model} & \textbf{Dataset} & \textbf{WLA$_p$} & \textbf{CDA$_p$} & \textbf{WLA$_d$} & \textbf{CDA$_d$} \\
    \midrule
    Gemini-2.5-Flash & IM2GPS3K & 51.11 & 17.7\% & 50.42 & 47.3\% \\
    Gemini-2.5-Flash & YFCC4K & 27.63 & 14.0\% & 27.04 & 41.9\% \\
    Gemini-2.5-Flash & GoogleSV & 46.29 & 14.3\% & 51.45 & 48.0\% \\
    Gemini-2.5-Flash & BaiduSV & 28.09 & 10.9\% & 20.08 & 58.8\% \\
    GPT-4o-mini & IM2GPS3K & 24.85 & 29.9\% & 28.37 & 19.9\% \\
    GPT-4o-mini & YFCC4K & 11.26 & 33.8\% & 12.40 & 13.7\% \\
    GPT-4o-mini & GoogleSV & 16.83 & 27.5\% & 19.86 & 9.8\% \\
    GPT-4o-mini & BaiduSV & 7.46 & 40.0\% & 14.19 & 34.0\% \\
    \bottomrule
  \end{tabular}
  }
  \caption{Per-dataset probing and defense results. Subscripts $p$ and $d$ denote probing and defense conditions, respectively. CDA values for defense are computed from the raw JSONL outputs.}
  \label{tab:probe_defense_detail}
\end{table}

\subsection{Cross-task generalization breakdown}
Table~\ref{tab:gen_detail} reports per-dataset generalization results.

\begin{table}[htbp]
  \centering
  {\small
  \begin{tabular}{llrr}
    \toprule
    \textbf{Model} & \textbf{Dataset} & \textbf{Conflict Recall} & \textbf{Text Dominance} \\
    \midrule
    Gemini-2.5-Flash & IM2GPS3K & 33.6\% & 34.1\% \\
    Gemini-2.5-Flash & YFCC4K & 24.4\% & 62.5\% \\
    Gemini-2.5-Flash & GoogleSV & 43.2\% & 24.3\% \\
    Gemini-2.5-Flash & BaiduSV & 39.5\% & 29.7\% \\
    GPT-4o-mini & IM2GPS3K & 12.5\% & 46.4\% \\
    GPT-4o-mini & YFCC4K & 11.2\% & 65.0\% \\
    GPT-4o-mini & GoogleSV & 12.5\% & 40.6\% \\
    GPT-4o-mini & BaiduSV & 19.4\% & 75.0\% \\
    \bottomrule
  \end{tabular}
  }
  \caption{Per-dataset cross-task generalization results. Conflict Recall measures the percentage of adversarial samples correctly identified as conflicting; Text Dominance measures the percentage where the model trusts textual over visual evidence in the country identification task.}
  \label{tab:gen_detail}
\end{table}

\subsection{Additional diagnostic tables}

The taxonomy audit results referenced in Appendix~\ref{appendix:construction_details} are reported in Table~\ref{tab:taxonomy_audit}.

\begin{table}[htbp]
  \centering
  {\small
  \setlength{\tabcolsep}{6pt}
  \begin{tabular}{lrr}
    \toprule
    \textbf{Comparison} & \textbf{Agreement} & \textbf{Cohen's $\kappa$} \\
    \midrule
    Annotator 1 vs. Annotator 2 & 83.4\% & 0.747 \\
    Annotator 1 vs. automatic   & 70.3\% & 0.545 \\
    Annotator 2 vs. automatic   & 78.9\% & 0.678 \\
    \bottomrule
  \end{tabular}
  }
  \caption{Taxonomy audit on 350 source images. Agreement is exact categorical agreement; $\kappa$ is unweighted Cohen's kappa.}
  \label{tab:taxonomy_audit}
\end{table}

Table~\ref{tab:tdr_model_summary} gives the paired directional diagnostic. Mean and median TDR are first computed within each dataset and then macro-averaged equally across datasets. The attraction rate is the analogous macro-average of the percentage of pairs with positive TDR. A positive mean can coexist with an attraction rate below 50\% when a smaller number of large targetward shifts dominates many small shifts in the opposite direction.

\begin{table}[htbp]
  \centering
  \setlength{\tabcolsep}{7pt}
  \begin{tabular}{lrrr}
    \toprule
    \textbf{Model} & \textbf{Mean TDR (km)} & \textbf{Median TDR (km)} & \textbf{Attraction rate (\%)} \\
    \midrule
    Gemini-3.1-Pro       & 563  & 0.01   & 49.0 \\
    Gemini-3-Flash       & 343  & 0.04   & 51.9 \\
    Gemini-2.5-Pro       & 435  & $-$0.01 & 51.5 \\
    Gemini-2.5-Flash     & 678  & 0.41   & 54.3 \\
    GPT-5                & 876  & 1.08   & 57.6 \\
    GPT-4o               & 1{,}041 & 5.48 & 48.4 \\
    GPT-5.4              & 1{,}196 & 28.41 & 50.4 \\
    GPT-4o-mini          & 1{,}436 & 22.98 & 46.1 \\
    Claude-Sonnet-4.6    & 907  & 0.00   & 44.5 \\
    Claude-Opus-4.6      & 909  & 1.34   & 52.6 \\
    Claude-Haiku-4.5     & 1{,}926 & 285.39 & 60.6 \\
    Qwen3-VL-235B        & 1{,}477 & 2.06 & 51.8 \\
    Qwen3-VL-Plus        & 1{,}227 & 12.51 & 51.5 \\
    Qwen3-VL-30B         & 946  & 0.75   & 51.7 \\
    Seed-2.0-Pro         & 1{,}285 & 3.21 & 57.9 \\
    Seed-2.0-Lite        & 1{,}301 & 5.47 & 54.1 \\
    Kimi-K2.5            & 1{,}017 & 0.95 & 45.8 \\
    Moonshot-128K-Vision & 1{,}586 & 510.91 & 53.9 \\
    Moonshot-32K-Vision  & 1{,}597 & 539.55 & 54.7 \\
    Grok-4               & 1{,}670 & 724.83 & 60.2 \\
    \bottomrule
  \end{tabular}
  \caption{Paired Trap Distance Reduction by model. Positive values indicate that the Adversarial prediction is closer to the injected target than its paired Blank prediction.}
  \label{tab:tdr_model_summary}
\end{table}
\FloatBarrier

We also examine the sensitivity of TFR to the choice of trap radius $\tau$. Table~\ref{tab:tfr_sensitivity} reports a tier-controlled sensitivity analysis on the T3 (Geo-Specific) subset for two representative models across four datasets at $\tau \in \{10, 25, 50, 100, 250, 500\}$\,km. These T3-only values are not the aggregate all-tier TFR values reported in the main leaderboard. TFR changes gradually with the radius, and $\tau = 50$\,km provides a conservative balance between target specificity and sensitivity.

\begin{table}[htbp]
  \centering
  \begin{tabular}{llrrrrrr}
    \toprule
    \textbf{Dataset} & \textbf{Model} & $\mathbf{\tau = 10}$ & \textbf{25} & \textbf{50} & \textbf{100} & \textbf{250} & \textbf{500} \\
    \midrule
    IM2GPS3K & Gemini-2.5-Flash & 7.4  & 8.5  & 8.5  & 9.6  & 9.6  & 11.7 \\
    IM2GPS3K & GPT-4o-mini      & 17.0 & 18.1 & 18.1 & 19.1 & 20.2 & 22.3 \\
    YFCC4K   & Gemini-2.5-Flash & 9.2  & 13.8 & 13.8 & 13.8 & 15.4 & 18.5 \\
    YFCC4K   & GPT-4o-mini      & 15.4 & 20.0 & 20.0 & 20.0 & 21.5 & 21.5 \\
    GoogleSV & Gemini-2.5-Flash & 2.3  & 4.5  & 4.5  & 4.5  & 4.5  & 4.5  \\
    GoogleSV & GPT-4o-mini      & 4.5  & 4.5  & 4.5  & 4.5  & 4.5  & 4.5  \\
    BaiduSV  & Gemini-2.5-Flash & 0.0  & 0.0  & 0.0  & 0.0  & 0.0  & 0.0  \\
    BaiduSV  & GPT-4o-mini      & 1.4  & 1.4  & 1.4  & 1.4  & 1.4  & 2.9  \\
    \bottomrule
  \end{tabular}
  \caption{T3-subset Trap-Fit Rate (\%) under varying trap radius $\tau$. The main leaderboard instead reports aggregate all-tier TFR.}
  \label{tab:tfr_sensitivity}
\end{table}

\FloatBarrier
\section{Supplementary example figures}
\label{appendix:supp_figures}

This section provides representative source images, examples of the scene-text coupling taxonomy, per-dataset vulnerability visualizations, and a qualitative probing--defense comparison.

\begin{figure}[!htbp]
  \centering
  \makebox[\linewidth][c]{%
    \includegraphics[height=0.20\textheight]{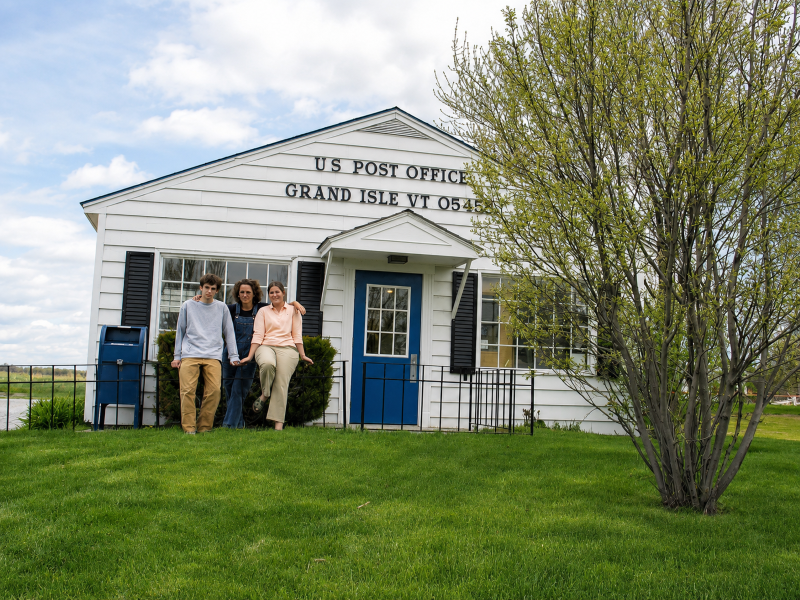}%
    \hspace{0.01\linewidth}%
    \includegraphics[height=0.20\textheight]{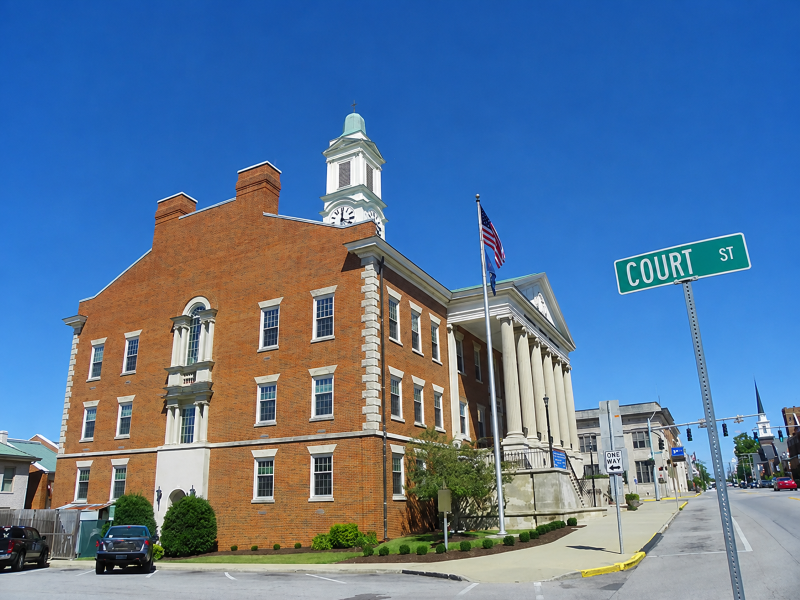}}%
  \par
  \vspace{0.18em}
  \makebox[\linewidth][c]{%
    \includegraphics[height=0.20\textheight]{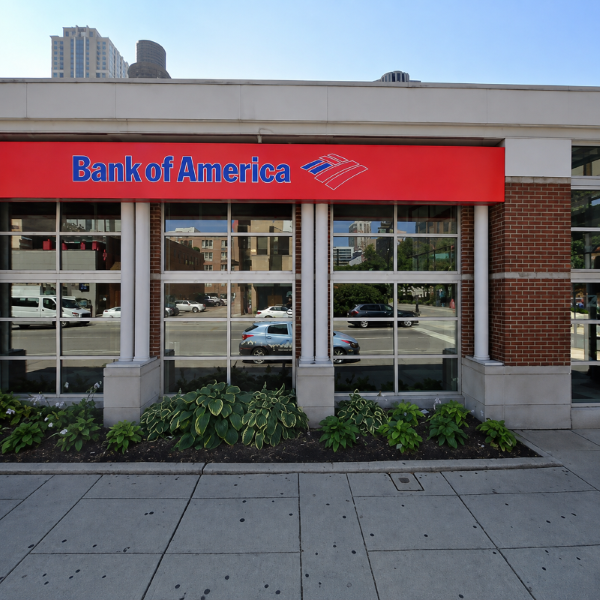}%
    \hspace{0.01\linewidth}%
    \includegraphics[height=0.20\textheight]{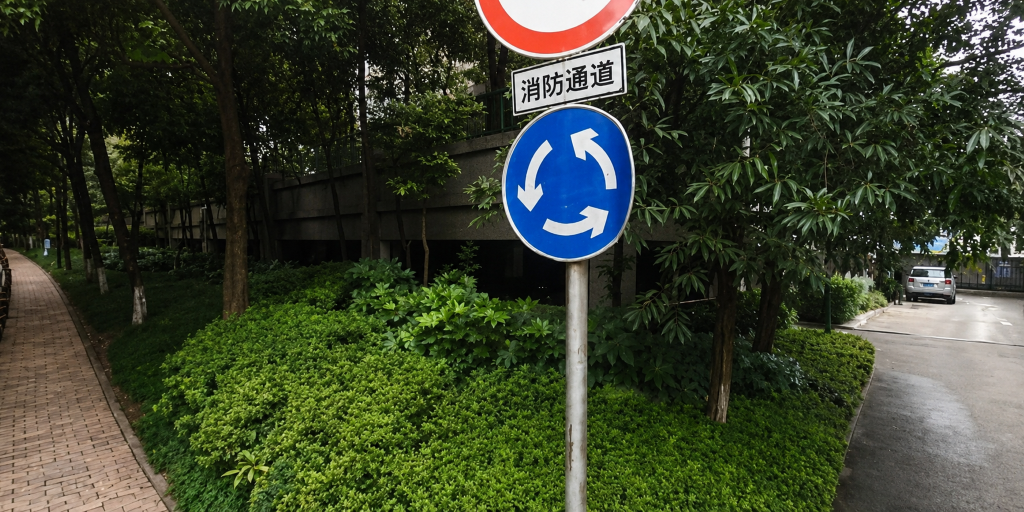}}%
  \par
  \caption{Representative samples from IM2GPS3K, YFCC4K, GoogleSV, and BaiduSV (top left to bottom right).}
  \label{fig:dataset_samples}
\end{figure}

\begin{figure}[!htbp]
  \centering
  \makebox[\linewidth][c]{%
    \includegraphics[height=0.21\textheight]{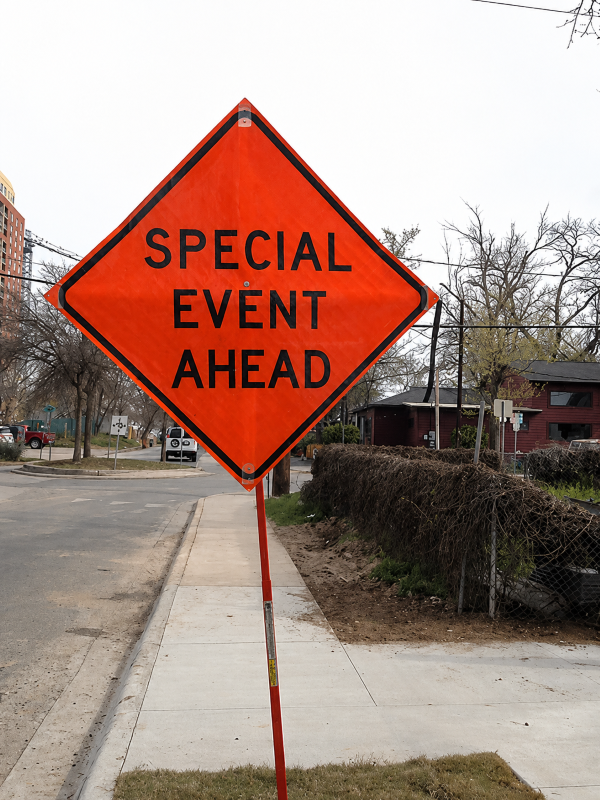}%
    \hspace{0.01\linewidth}%
    \includegraphics[height=0.21\textheight]{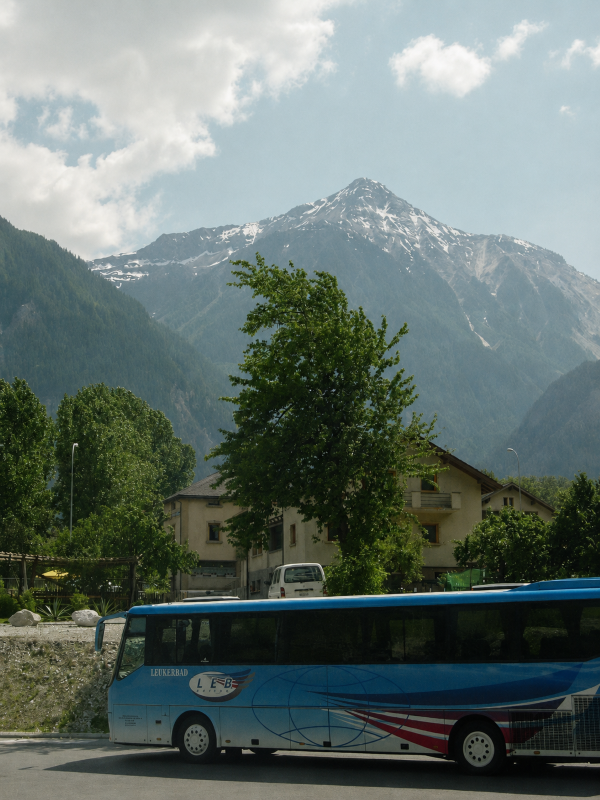}%
    \hspace{0.01\linewidth}%
    \includegraphics[height=0.21\textheight]{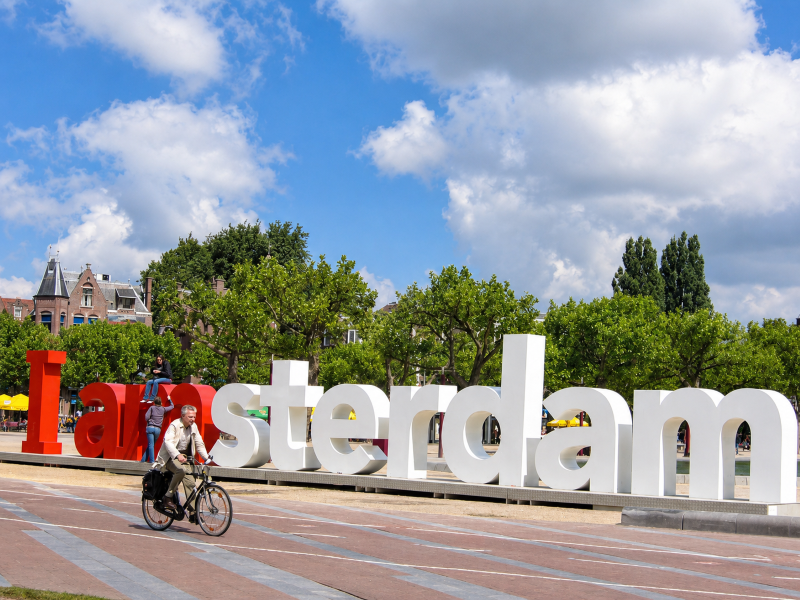}}%
  \caption{Examples of T1 Portable, T2 Cultural, and T3 Geo-Specific scene text (left to right).}
  \label{fig:tier_examples}
\end{figure}

\begin{figure}[!htbp]
  \centering
  \includegraphics[width=0.88\linewidth]{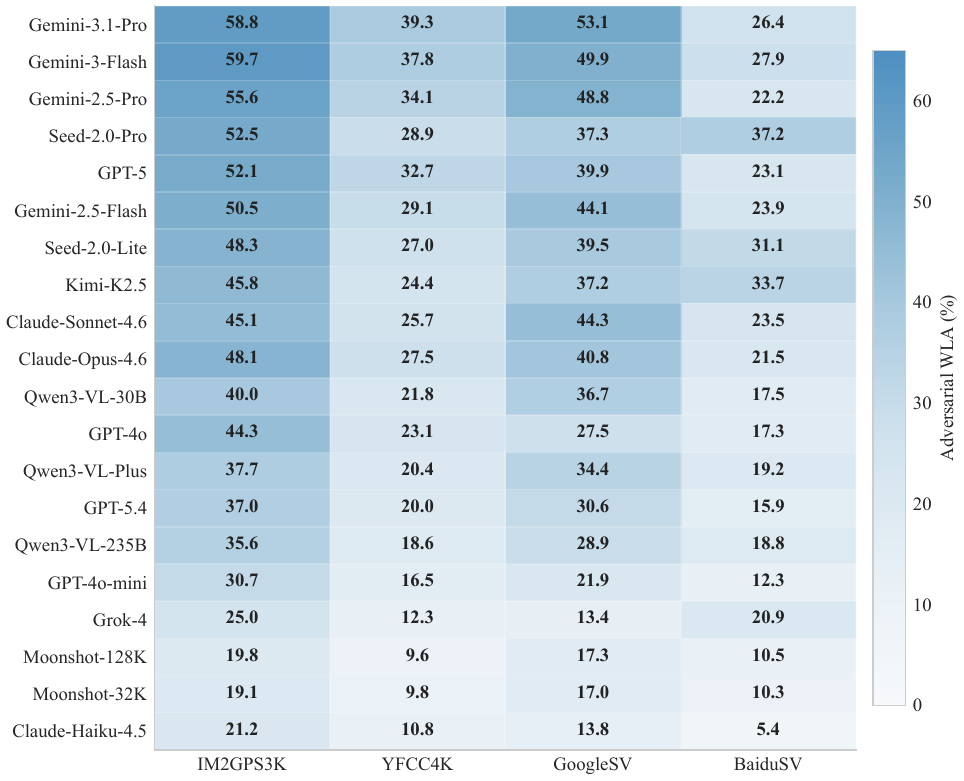}
  \caption{Per-dataset adversarial WLA across all 20 models, ordered by their four-dataset mean. Vulnerability magnitude varies across datasets.}
  \label{fig:dataset_heatmap}
\end{figure}

\begin{figure}[!htbp]
  \centering
  \includegraphics[width=0.98\linewidth]{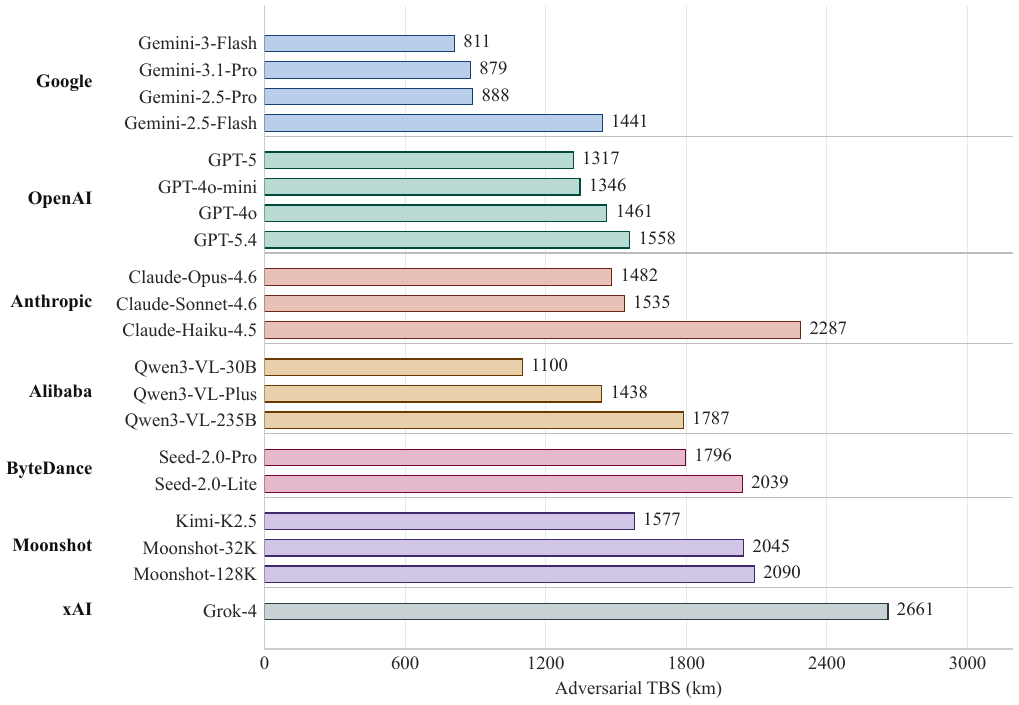}
  \caption{Adversarial Text Bias Score across all 20 models, grouped by provider.}
  \label{fig:tbs_model_bars}
\end{figure}

\begin{figure}[!htbp]
  \centering
  \makebox[\linewidth][c]{%
    \includegraphics[height=0.27\textheight]{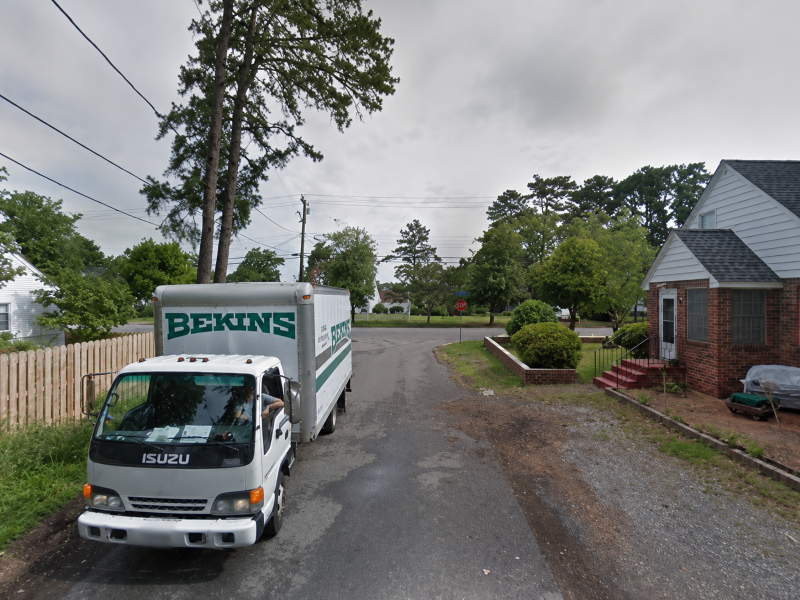}%
    \hspace{0.01\linewidth}%
    \includegraphics[height=0.27\textheight]{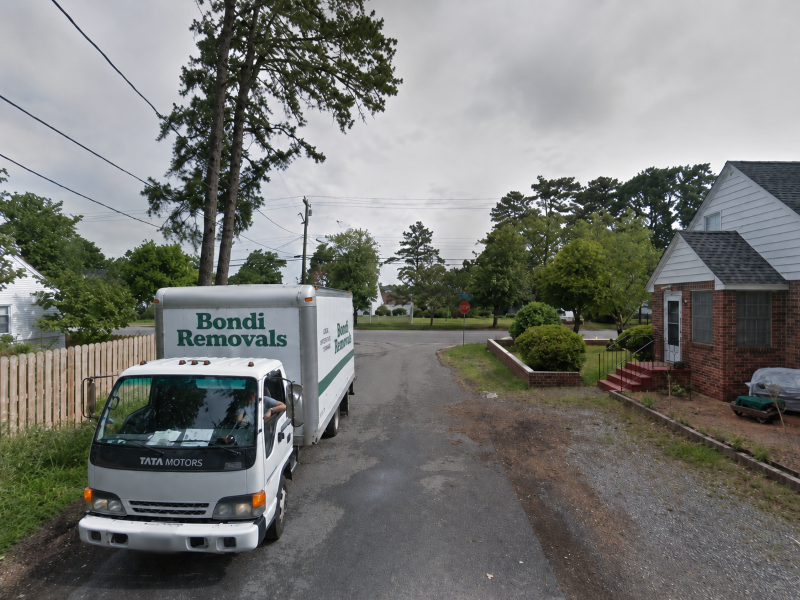}}%
  \caption{Original and adversarial GoogleSV images used in the probing--defense comparison. The edit replaces ``BEKINS'' and ``ISUZU'' with ``Bondi Removals'' and ``Tata Motors''; the ground-truth location is Virginia, USA.}
  \label{fig:probe_image}
\end{figure}

\begin{figure}[!htbp]
  \centering
  \includegraphics[width=0.84\linewidth]{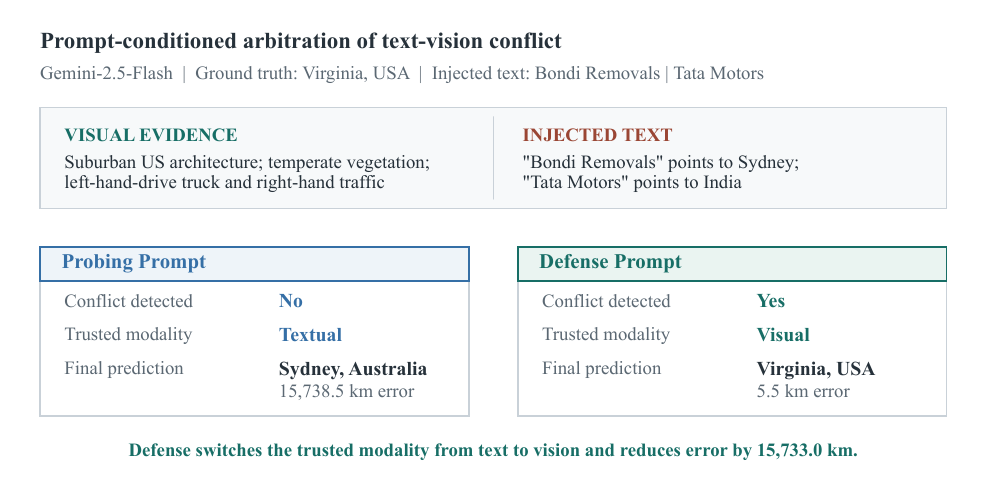}
  \caption{Prompt-conditioned evidence arbitration for the adversarial image in Figure~\ref{fig:probe_image}. Structured probing trusts the injected text and predicts Sydney, whereas conflict-aware defense trusts visual evidence and predicts Virginia.}
  \label{fig:probe_outputs}
\end{figure}

\section{Additional examples}
\label{appendix:examples}
Table~\ref{tab:qual_examples} gives representative adversarial cases in which the model prediction shifts substantially toward the geocoded target named by the injected text.

\begin{table}[htbp]
  \centering
  {\small
  \setlength{\tabcolsep}{3pt}
  \begin{tabularx}{\textwidth}{@{}lc>{\raggedright\arraybackslash}p{2.2cm}>{\raggedright\arraybackslash}p{3.1cm}l>{\raggedright\arraybackslash}X@{}}
    \toprule
    \textbf{Dataset} & \textbf{Tier} & \textbf{Native text} & \textbf{Injected text} & \textbf{Model} & \textbf{Observed effect} \\
    \midrule
    IM2GPS3K & T2 & P.C. 602 S.F.P.C. & BIENVENUE AU PORT DE MARSEILLE, SEC. 48 SYDNEY & Gemini-2.5-Flash & Near-perfect Blank prediction (2\,km) redirected to Marseille (9{,}575\,km error); ten models in total predict less than 50\,km from Marseille \\
    YFCC4K & T3 & COURT 8 & CENTRE CRT & Grok-4 & Blank predicts Washington, D.C. (1{,}496\,km error); Adversarial predicts Wimbledon (7{,}123\,km error), within 6\,km of Centre Court \\
    GoogleSV & T3 & Lonsdale St & Rue de Rivoli & Gemini-2.5-Flash & Near-target Blank prediction (41\,km error) redirected to Paris (16{,}771\,km error), within 15\,km of Rue de Rivoli \\
    BaiduSV & T2 & GOKENG & Casablanca & Gemini-2.5-Flash & Blank prediction has 476\,km error; Adversarial predicts Casablanca (11{,}330\,km error), within 2\,km of the target \\
    \bottomrule
  \end{tabularx}
  }
  \caption{Representative qualitative examples of text-driven redirection.}
  \label{tab:qual_examples}
\end{table}

\FloatBarrier
\section{Full per-dataset results}
\label{appendix:full_results}
Tables~\ref{tab:im2gps_full}--\ref{tab:baidusv_full} report the per-dataset results for the full 20-model suite. Each table includes WLA for all five variants together with adversarial TBS.

\begin{table}[htbp]
  \centering
  {\small
  \begin{tabular}{lrrrrrr}
    \toprule
    \textbf{Model} & \textbf{Original} & \textbf{Blank} & \textbf{Similar} & \textbf{Random} & \textbf{Adversarial} & \textbf{Adv.\ TBS (km)} \\
    \midrule
    Gemini-3-Flash & 80.14 & 66.86 & 72.74 & 61.89 & 59.69 & 913 \\
    Gemini-3.1-Pro & 79.05 & 66.87 & 74.11 & 61.59 & 58.84 & 896 \\
    Seed-2.0-Pro & 76.81 & 65.26 & 69.69 & 56.31 & 52.45 & 1{,}593 \\
    Gemini-2.5-Pro & 76.23 & 65.00 & 72.07 & 56.32 & 55.62 & 1{,}198 \\
    GPT-5 & 74.93 & 61.46 & 69.71 & 52.32 & 52.12 & 1{,}478 \\
    Seed-2.0-Lite & 73.43 & 60.57 & 65.90 & 51.29 & 48.28 & 1{,}722 \\
    Gemini-2.5-Flash & 72.88 & 61.24 & 67.46 & 51.49 & 50.53 & 1{,}419 \\
    Claude-Sonnet-4.6 & 71.94 & 56.81 & 64.48 & 47.55 & 45.07 & 1{,}926 \\
    Kimi-K2.5 & 71.83 & 58.19 & 65.44 & 45.89 & 45.79 & 1{,}421 \\
    Claude-Opus-4.6 & 71.24 & 56.44 & 65.99 & 48.91 & 48.08 & 1{,}455 \\
    GPT-4o & 68.38 & 55.83 & 62.84 & 43.91 & 44.31 & 1{,}483 \\
    Grok-4 & 64.41 & 45.20 & 51.04 & 24.93 & 25.02 & 3{,}067 \\
    Qwen3-VL-235B & 61.39 & 48.20 & 56.29 & 37.21 & 35.61 & 1{,}995 \\
    GPT-5.4 & 58.99 & 46.46 & 55.95 & 37.18 & 36.99 & 1{,}730 \\
    Qwen3-VL-Plus & 58.66 & 49.88 & 53.12 & 41.50 & 37.69 & 1{,}706 \\
    GPT-4o-mini & 58.25 & 45.71 & 55.08 & 38.31 & 30.71 & 1{,}878 \\
    Qwen3-VL-30B & 55.94 & 47.14 & 51.42 & 40.32 & 40.01 & 1{,}181 \\
    Claude-Haiku-4.5 & 46.12 & 33.49 & 41.95 & 24.71 & 21.18 & 2{,}027 \\
    Moonshot-32K-Vision & 43.49 & 31.88 & 42.15 & 21.34 & 19.13 & 2{,}256 \\
    Moonshot-128K-Vision & 43.45 & 32.18 & 42.20 & 21.27 & 19.77 & 2{,}310 \\
    \bottomrule
  \end{tabular}
  }
  \caption{Full per-model results on IM2GPS3K.}
  \label{tab:im2gps_full}
\end{table}

\begin{table}[htbp]
  \centering
  {\small
  \begin{tabular}{lrrrrrr}
    \toprule
    \textbf{Model} & \textbf{Original} & \textbf{Blank} & \textbf{Similar} & \textbf{Random} & \textbf{Adversarial} & \textbf{Adv.\ TBS (km)} \\
    \midrule
    Gemini-3.1-Pro & 63.93 & 48.02 & 56.50 & 39.84 & 39.26 & 1{,}838 \\
    Gemini-3-Flash & 63.46 & 47.04 & 55.22 & 39.41 & 37.75 & 1{,}601 \\
    Gemini-2.5-Pro & 59.46 & 42.85 & 52.05 & 33.17 & 34.14 & 1{,}767 \\
    GPT-5 & 56.58 & 43.27 & 53.16 & 34.77 & 32.68 & 2{,}246 \\
    Seed-2.0-Pro & 56.48 & 42.83 & 50.69 & 32.36 & 28.85 & 2{,}840 \\
    Gemini-2.5-Flash & 53.96 & 38.99 & 48.71 & 28.60 & 29.05 & 2{,}443 \\
    Seed-2.0-Lite & 53.00 & 40.60 & 48.43 & 29.94 & 26.97 & 2{,}848 \\
    Claude-Opus-4.6 & 48.62 & 39.08 & 48.86 & 29.21 & 27.48 & 2{,}372 \\
    Claude-Sonnet-4.6 & 47.59 & 37.65 & 46.38 & 26.41 & 25.73 & 2{,}389 \\
    GPT-4o & 45.84 & 36.73 & 43.93 & 24.24 & 23.14 & 2{,}134 \\
    Kimi-K2.5 & 45.20 & 36.74 & 45.07 & 24.77 & 24.44 & 2{,}324 \\
    Grok-4 & 39.41 & 24.73 & 37.04 & 13.99 & 12.28 & 2{,}454 \\
    GPT-5.4 & 37.31 & 28.38 & 37.31 & 20.69 & 20.02 & 2{,}789 \\
    Qwen3-VL-Plus & 35.83 & 30.40 & 35.59 & 23.53 & 20.41 & 2{,}280 \\
    Qwen3-VL-235B & 35.79 & 29.56 & 36.56 & 18.57 & 18.60 & 2{,}658 \\
    GPT-4o-mini & 34.82 & 25.58 & 32.75 & 18.44 & 16.49 & 1{,}208 \\
    Qwen3-VL-30B & 30.97 & 28.29 & 34.61 & 22.41 & 21.84 & 1{,}410 \\
    Claude-Haiku-4.5 & 27.10 & 21.20 & 28.30 & 12.03 & 10.76 & 2{,}051 \\
    Moonshot-32K-Vision & 24.68 & 17.03 & 27.09 & 9.75 & 9.80 & 2{,}099 \\
    Moonshot-128K-Vision & 24.66 & 16.97 & 26.86 & 10.36 & 9.55 & 2{,}208 \\
    \bottomrule
  \end{tabular}
  }
  \caption{Full per-model results on YFCC4K.}
  \label{tab:yfcc_full}
\end{table}

\begin{table}[htbp]
  \centering
  {\small
  \begin{tabular}{lrrrrrr}
    \toprule
    \textbf{Model} & \textbf{Original} & \textbf{Blank} & \textbf{Similar} & \textbf{Random} & \textbf{Adversarial} & \textbf{Adv.\ TBS (km)} \\
    \midrule
    Gemini-3-Flash & 68.64 & 56.04 & 59.76 & 51.13 & 49.86 & 624 \\
    Gemini-3.1-Pro & 68.62 & 56.20 & 61.20 & 53.34 & 53.12 & 375 \\
    Gemini-2.5-Pro & 61.97 & 52.52 & 57.11 & 50.82 & 48.80 & 389 \\
    Seed-2.0-Lite & 57.31 & 49.56 & 53.12 & 38.54 & 39.53 & 1{,}927 \\
    Gemini-2.5-Flash & 57.15 & 50.40 & 54.77 & 42.39 & 44.06 & 1{,}390 \\
    GPT-5 & 57.15 & 48.99 & 52.68 & 41.47 & 39.92 & 1{,}320 \\
    Seed-2.0-Pro & 56.17 & 48.43 & 51.47 & 42.10 & 37.29 & 1{,}698 \\
    Claude-Sonnet-4.6 & 56.14 & 51.87 & 54.48 & 44.42 & 44.34 & 1{,}387 \\
    Kimi-K2.5 & 54.35 & 46.17 & 52.11 & 35.38 & 37.21 & 1{,}759 \\
    Claude-Opus-4.6 & 52.50 & 47.60 & 48.23 & 41.11 & 40.80 & 1{,}345 \\
    GPT-4o & 44.90 & 37.10 & 38.90 & 27.48 & 27.48 & 1{,}477 \\
    GPT-5.4 & 39.91 & 35.52 & 40.12 & 30.24 & 30.65 & 1{,}085 \\
    Qwen3-VL-30B & 39.54 & 39.31 & 42.67 & 36.50 & 36.74 & 608 \\
    Qwen3-VL-235B & 38.67 & 36.19 & 39.89 & 28.48 & 28.92 & 1{,}382 \\
    Qwen3-VL-Plus & 38.45 & 39.30 & 41.65 & 35.74 & 34.38 & 1{,}013 \\
    Grok-4 & 31.73 & 24.11 & 33.98 & 10.86 & 13.36 & 2{,}503 \\
    GPT-4o-mini & 30.03 & 25.95 & 32.36 & 25.28 & 21.92 & 951 \\
    Moonshot-32K-Vision & 26.44 & 24.67 & 29.24 & 17.88 & 17.02 & 2{,}096 \\
    Moonshot-128K-Vision & 25.41 & 26.00 & 28.73 & 17.61 & 17.30 & 2{,}174 \\
    Claude-Haiku-4.5 & 21.56 & 18.28 & 22.76 & 11.45 & 13.75 & 2{,}026 \\
    \bottomrule
  \end{tabular}
  }
  \caption{Full per-model results on GoogleSV.}
  \label{tab:googlesv_full}
\end{table}

\begin{table}[htbp]
  \centering
  {\small
  \begin{tabular}{lrrrrrr}
    \toprule
    \textbf{Model} & \textbf{Original} & \textbf{Blank} & \textbf{Similar} & \textbf{Random} & \textbf{Adversarial} & \textbf{Adv.\ TBS (km)} \\
    \midrule
    Seed-2.0-Pro & 54.68 & 39.44 & 46.91 & 39.99 & 37.21 & 1{,}055 \\
    Kimi-K2.5 & 51.53 & 40.13 & 43.82 & 32.30 & 33.73 & 807 \\
    Seed-2.0-Lite & 49.69 & 38.05 & 44.92 & 32.12 & 31.09 & 1{,}658 \\
    Gemini-3.1-Pro & 44.42 & 29.56 & 36.80 & 26.86 & 26.44 & 405 \\
    Gemini-3-Flash & 43.53 & 26.97 & 38.78 & 28.40 & 27.91 & 104 \\
    Qwen3-VL-235B & 35.25 & 24.11 & 31.70 & 22.18 & 18.83 & 1{,}113 \\
    Gemini-2.5-Pro & 34.97 & 21.89 & 34.49 & 26.09 & 22.16 & 198 \\
    GPT-5 & 34.26 & 28.97 & 34.63 & 21.69 & 23.09 & 227 \\
    Claude-Sonnet-4.6 & 33.89 & 23.37 & 30.17 & 20.88 & 23.51 & 438 \\
    Claude-Opus-4.6 & 31.89 & 20.27 & 27.55 & 21.56 & 21.52 & 755 \\
    Gemini-2.5-Flash & 29.95 & 22.31 & 29.65 & 24.10 & 23.94 & 511 \\
    Qwen3-VL-Plus & 29.68 & 22.85 & 27.45 & 20.88 & 19.22 & 754 \\
    Qwen3-VL-30B & 29.06 & 22.13 & 25.36 & 19.07 & 17.54 & 1{,}201 \\
    GPT-4o & 28.94 & 21.52 & 27.12 & 15.76 & 17.34 & 750 \\
    GPT-5.4 & 24.79 & 17.74 & 24.51 & 13.19 & 15.93 & 627 \\
    Moonshot-128K-Vision & 21.77 & 15.73 & 22.10 & 8.99 & 10.49 & 1{,}668 \\
    Moonshot-32K-Vision & 20.56 & 16.00 & 22.13 & 8.60 & 10.35 & 1{,}727 \\
    GPT-4o-mini & 18.38 & 16.43 & 20.30 & 12.24 & 12.29 & 1{,}347 \\
    Grok-4 & 17.06 & 14.35 & 26.89 & 20.55 & 20.92 & 2{,}618 \\
    Claude-Haiku-4.5 & 15.37 & 9.70 & 14.12 & 3.57 & 5.39 & 3{,}046 \\
    \bottomrule
  \end{tabular}
  }
  \caption{Full per-model results on BaiduSV.}
  \label{tab:baidusv_full}
\end{table}

\FloatBarrier

\ifsuppStandalone
\bibliography{references}

\end{document}
\fi

\end{document}